\pdfoutput=1
\documentclass[lettersize,journal]{IEEEtran}
\usepackage{amsmath,amsfonts}
\usepackage{algorithmic}
\usepackage{algorithm}
\usepackage{array}
\usepackage[caption=false,font=normalsize,labelfont=sf,textfont=sf]{subfig}
\usepackage{textcomp}
\usepackage{stfloats}
\usepackage{url}
\usepackage{verbatim}
\usepackage{graphicx}
\usepackage{cite}
\usepackage[most]{tcolorbox}
\usepackage{booktabs}
\usepackage{multirow}
\usepackage{adjustbox} 
\usepackage{subcaption}
\usepackage{orcidlink}
\newcommand{\suporcid}[1]{\textsuperscript{\orcidlink{#1}}}
\usepackage{hyperref}
\hypersetup{
  colorlinks=true,
  allcolors=blue,
}

\begin{document}
\bstctlcite{IEEEexample:BSTcontrol}

\title{Multilingual Training-Free Remote Sensing Image Captioning}

\author{%
Carlos Rebelo\suporcid{0009-0004-6161-9429},
Gil Rocha\suporcid{0000-0001-8252-7292},
João Daniel Silva\suporcid{0000-0001-6474-7822},
Bruno Martins\suporcid{0000-0002-3856-2936}%
\thanks{The authors are with INESC-ID, Lisboa, Portugal. Carlos Rebelo (e-mail: carlosrebelo17@tecnico.ulisboa.pt), João Daniel Silva and Bruno Martins are also with Instituto Superior Técnico, Universidade de Lisboa, Portugal. Gil Rocha is also with Faculdade de Engenharia da Universidade do Porto, Portugal.}
}



\maketitle

\begin{abstract}
Remote sensing image captioning has advanced rapidly through encoder–decoder models, although the reliance on large annotated datasets and the focus on English restricts global applicability. To address these limitations, we propose the first training-free multilingual approach, based on retrieval-augmented prompting. For a given aerial image, we employ a domain-adapted SigLIP2 encoder to retrieve related captions and few-shot examples from a datastore, which are then provided to a language model. We explore two variants: an image-blind setup, where a multilingual Large Language Model (LLM) generates the caption from textual prompts alone, and an image-aware setup, where a Vision–Language Model (VLM) jointly processes the prompt and the input image. To improve the coherence of the retrieved content, we introduce a graph-based re-ranking strategy using PageRank on a graph of images and captions. Experiments on four benchmark datasets across ten languages demonstrate that our approach is competitive with fully supervised English-only systems and generalizes to other languages. Results also highlight the importance of re-ranking with PageRank, yielding up to 35\% improvements in performance metrics. Additionally, it was observed that while VLMs tend to generate visually grounded but lexically diverse captions, LLMs can achieve stronger BLEU and CIDEr scores. Lastly, directly generating captions in the target language consistently outperforms other translation-based strategies. Overall, our work delivers one of the first systematic evaluations of multilingual, training-free captioning for remote sensing imagery, advancing toward more inclusive and scalable multimodal Earth observation systems.
\end{abstract}


\section{Introduction}

\IEEEPARstart{R}{emote} sensing image captioning, i.e. the task of generating textual descriptions for aerial images, has seen significant progress through encoder-decoder models. However, the reliance on large annotated training datasets, and the prevailing focus on the English language, limits the accessibility and applicability of this technology for a global user base.

Instead of adhering to the conventional supervised training paradigm, we propose a training-free approach for multilingual captioning that does not require manually annotated data for model training. Our method instead hinges on prompting a multilingual Large Language Model (LLM) with captions from similar images to the input image. This aligns with recent findings demonstrating that retrieval-based prompting can achieve state-of-the-art performance with LLMs, bypassing the need for large-scale multimodal training data~\cite{tewel2022zerocap,ramos2023lmcap,kim2024nice}.

Our primary contribution is the application and analysis of a few-shot methodology within the remote sensing domain and in a multilingual setting. Specifically, for a given aerial image, we employ a SigLIP2 encoder~\cite{tschannen2025siglip2}, fine-tuned for the remote sensing domain, to retrieve relevant captions and similar images, which also have relevant captions associated, forming few-shot examples. All these captions are then consolidated into a prompt to guide the language model in generating the final description. We explore two variants of this approach, namely an image-blind version that provides only the textual prompt to a Large Language Model (LLM), and an image-aware version that supplies both the prompt and the input image to a Vision-Language Model (VLM).

To enhance the quality and coherence of the information used in the prompt, we also introduce a novel graph-based re-ranking strategy. This method leverages the personalized PageRank algorithm~\cite{Page1998PageRank,Gleich2015PageRank} on a multimodal graph where retrieved images and captions are represented as nodes. Node and edge weights are determined by their similarity to the target image and to each other, computed using the same visual encoder from the retrieval stage. This process re-ranks the retrieved elements to better capture their collective relevance, which we hypothesize to lead to more effective prompts.

Experiments on established benchmark datasets demonstrate that our approach is competitive with fully supervised English-only models, all while enabling multilingual captioning without any dedicated training. Furthermore, our analysis confirms that the graph-based re-ranking strategy yields significant improvements, by enhancing the semantic coherence of the elements within the prompt.

In summary, we propose a multilingual remote sensing image captioning method that leverages cross-modal retrieval to prompt large-scale language and vision-language models. The main contributions of this work are as follows:

\begin{itemize}
\item We demonstrate that our approaches, both image-blind (using LLMs) and image-aware (using VLMs), achieve results that are competitive with fully-supervised methods for English remote sensing image captioning.
\item We provide one of the first systematic evaluations of remote sensing image captioning in a multilingual setting, showing consistent performance across languages.
\item We propose a graph-based re-ranking strategy to improve the coherence of retrieved information for prompting, and we experimentally validate its effectiveness.
\end{itemize}

The rest of this paper is organized as follows: Section~\ref{sec:rwork} surveys related work, while Section~\ref{sec:method} describes the proposed approach. Section~\ref{sec:exp_setup} details the experimental setup, including the considered datasets and evaluation metrics, and Section~\ref{sec:results} reports the obtained results. Finally, Section~\ref{sec:conclusion} summarizes the main conclusions and discusses directions for future work.

\section{Related Work}
\label{sec:rwork}
The interaction between natural language and images is nowadays a popular topic within the machine learning, natural language processing, and computer vision communities, where researchers have worked on problems such as cross-modal retrieval, visual question answering, referring expression segmentation, or image captioning. Addressing these problems is particularly relevant in the context of remote sensing~\cite{bashmal2023language}, given that the current volume of remotely sensed Earth Observation (EO) data clearly motivates the development of vision-and-language methods that can support natural language interactions with large EO image repositories. 

Models based on Contrastive Language-Image Pre-training (CLIP) and Sigmoid Language-Image Pre-training (SigLIP) are nowadays extensively used in a variety of vision-and-language tasks~\cite{radfordICML2021learning, tschannen2025siglip2}, e.g. as dual encoders that can support cross-modal retrieval, or as vision encoders within larger models. In brief, CLIP models are based on an architecture with two separate encoders, one for visual inputs and one for textual inputs, trained on large-scale datasets with a contrastive objective that encourages semantically similar concepts, across both modalities, to be aligned in the shared representation space. The main difference between CLIP and SigLIP lies in the training objective: while CLIP employs a contrastive loss, SigLIP uses a sigmoid-based loss that treats each image–text pair independently. Despite this distinction, both approaches share the same overall architecture and the aim of learning a joint embedding space. Motivated by the strong results in tasks involving generalist images, the adaptation of CLIP to specific domains is currently being actively researched. In the specific domain of remote sensing, recent efforts include RemoteCLIP~\cite{remoteclip} or RS-M-CLIP~\cite{silva2024multilingual}, achieving very strong results in tasks such as cross-modal retrieval or zero-shot image classification.

CLIP vision encoders are also commonly used as backbones within recent architectures leveraging LLMs for vision-and-language tasks. For instance, LLaVA~\cite{liu2023visual}, MiniGPT-v2~\cite{chen2023minigpt}, or InstructBLIP~\cite{dai2023instructblip} are examples of instruction fine-tuned models supporting conversation and reasoning over visual inputs, connecting a CLIP vision encoder to a pre-trained LLM, and using supervised learning to train the cross-modal connections. Considering the remote sensing domain, similar efforts include RSGPT~\cite{hu2023rsgpt}, GeoChat~\cite{kuckreja2023geochat}, SkyEyeGPT~\cite{zhan2024skyeyegpt}, LHRS-Bot~\cite{muhtar2024lhrs}, or TEOChat~\cite{irvin24teochat}. Models like these address not only image captioning, which is the focus of the present study, but also other multimodal tasks (e.g., visual question answering) through a common interface based on prompting the models with task-specific instructions. However, specialized encoder-decoder models for image captioning correspond to the current state-of-the-art~\cite{ramos2022Usinga}, in some cases combining image encoders based on CLIP together with modules specifically designed to handle the properties of remote sensing imagery. For instance, models like MLCA-Net~\cite{cheng2022NWPUCaptions}, HCNet~\cite{yang2024hcnet}, MC-Net~\cite{huang2023mc}, BITA~\cite{yang2024bootstrapping}, the Aware-Transformer~\cite{cao2023aware}, or the Deformable Transformer~\cite{du2023deformable} all contain specialized modules to adaptively aggregate image features of multiple scales and specific regions. 

Despite the strong empirical results, all the aforementioned models depend on large datasets pairing aerial images to captions for supervised training. Extending these methods to multilingual settings is particularly challenging, requiring annotated data covering each target language, and larger models to mitigate the curse of multilinguality. Some previous studies have alternatively proposed training-free strategies for connecting vision encoders to LLMs. For instance, Socratic models~\cite{zeng2022socratic} correspond to a generalist framework in which an LLM is used to process a prompt that integrates information from different models. Under this framework, a task like image captioning can be performed using a CLIP model to retrieve different concepts from an input image (e.g., locations or objects that are likely featured in the image), and then prompting an LLM with the retrieved information. More recently, building on earlier approaches for retrieval-augmented generation of image captions~\cite{ramos2023retrieval,ramos2022smallcap}, Language Model Prompt-based Captioning (LMCap) considered a similar strategy~\cite{ramos2023lmcap}, addressing multilingual image captioning through the use of CLIP to retrieve similar captions from a datastore of examples, which are then used to prompt an LLM into generating the target caption. Experimental results showed that LMCap largely outperformed a Socratic modeling approach and performed competitively against supervised approaches for multilingual image captioning, without requiring expensive training with large-scale multimodal data. 

This paper essentially describes the application of LMCap to the domain of remote sensing imagery, using a SigLIP2 visual encoder and at the same time extending this approach in several directions. The extensions include a graph-based re-ranking strategy with personalized PageRank~\cite{Page1998PageRank,Gleich2015PageRank}, and a new image-aware approach, that uses VLMs instead of LLMs, and with the input image being prompted together with the retrieved information. The following section details the general procedure and also the proposed extensions.

\section{Proposed Approach}
\label{sec:method}
This section introduces our multilingual training-free captioning framework for remote sensing imagery, designed around retrieval-augmented generation. The framework can be used in two complementary modes: an image-blind strategy, where captions are produced without directly encoding the input image, and an image-aware strategy, where the input image is also provided to a frozen vision–language model. Both modes rely on a common retrieval and re-ranking backbone, diverging only at the final stage, when prompting the generative model.

\begin{figure*}
  \centering
  \includegraphics[width=\textwidth]{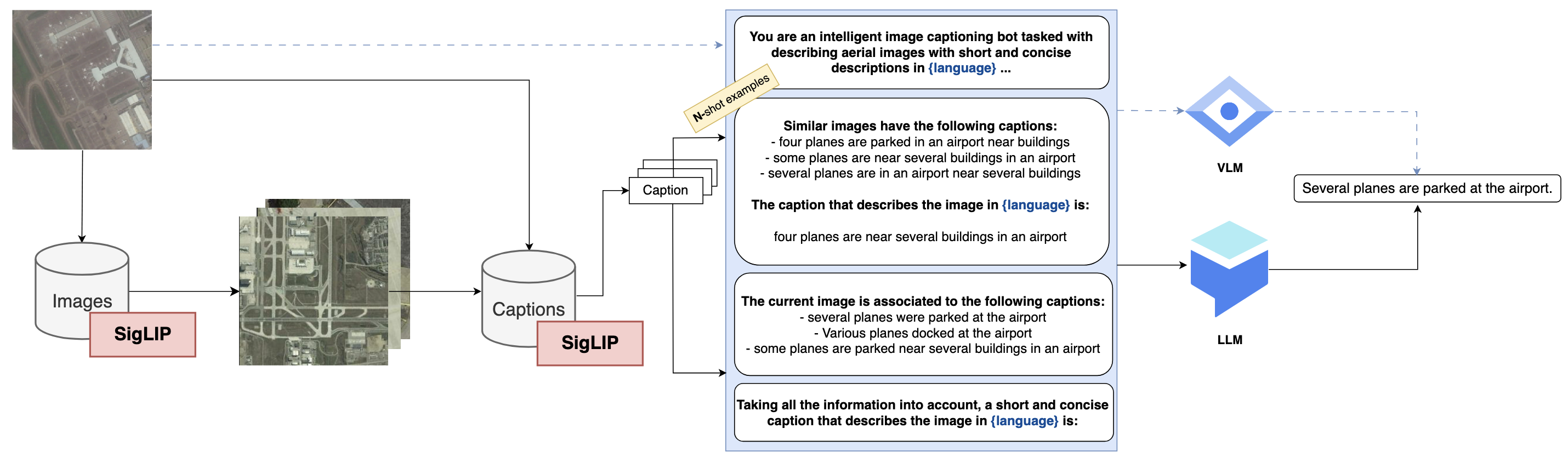}
  \caption{A general illustration for the proposed image captioning approach. For a given input image, the system first retrieves related captions, plus the most similar images and, for each of the images, also the corresponding most similar captions. The retrieved information is then assembled into a prompt, which can either be provided alone to a multilingual LLM (solid line) or, together with the input image, to a multilingual VLM (dashed line). In both cases, the model generates a caption in the target language. The prompt illustrated in the figure is a simplified example.}
  \label{fig:proposed_approach_scheme}
\end{figure*}

The process starts by retrieving semantically related captions from a large corpus using a domain-adapted SigLIP2 encoder (Section~\ref{section:ret_similar_capt}). The same encoder is then employed to retrieve visually similar images, whose captions serve as few-shot demonstrations of the captioning task (Section~\ref{section:few_shot_examples}). For each of the similar images, additional captions are collected along with their gold-standard annotations, in order to build the demonstrations. To enhance both coherence and diversity, the retrieved material is reorganized through a graph-based re-ranking procedure that applies personalized PageRank to emphasize semantically central items (Section~\ref{section:rerank_pagerank}). Finally, the re-ranked examples and captions are arranged into a prompt (Section~\ref{section:llm_prompting}). In the image-blind mode, this prompt is given to a multilingual LLM, which generates the caption without visual access. In the image-aware mode, the prompt is processed by a VLM that jointly considers the image and the retrieved textual context. An overview of both modes for the proposed pipeline is shown in Figure~\ref{fig:proposed_approach_scheme}.

\subsection{Retrieving Similar Captions}
\label{section:ret_similar_capt}

The proposed approach begins by retrieving captions that are semantically related to the input image, which are later used to construct a prompt for the language model. This retrieval step is based on version of the SigLIP2~\cite{tschannen2025siglip2} model that was fine-tuned for the remote sensing domain, here called RS-SigLIP2. This model was fine-tuned in previous work using the same procedure as RS-M-CLIP~\cite{silva2024multilingual}. Similarly to other CLIP-style models, RS-SigLIP2 incorporates aligned image and text encoders, denoted by $f_\phi$ and $g_\phi$, which map inputs into a shared multimodal embedding space.

First, a datastore $\mathcal{D} = \{\mathbf{\hat{x}}_i, \mathbf{\hat{y}}_i\}_{i=1}^{M}$ is built from a large set of remote sensing images, paired to captions in the English language, with $M$ image-caption pairs. 
The datastore indexes vector representations produced by RS-SigLIP2 for each image $f_\phi(\mathbf{\hat{x}}_i) = \mathbf{\hat{v}}_i \in \mathbb{R}^{d}$ and each caption $g_\phi(\mathbf{\hat{y}}_i) = \mathbf{\hat{u}}_i \in \mathbb{R}^{d}$.

At inference time, the input image is also encoded with the RS-SigLIP2 model (i.e., $f_\phi(\mathbf{x}_z) = \mathbf{v}_z$), and the image representation is used to retrieve the most similar captions. The similarity is obtained by calculating the cosine similarity between the vector representation of the image and each caption $i$ on the datastore $\mathcal{D}$, expressed by $\text{sim}(\mathbf{x_z}, \mathbf{\hat{y}_i})=\cos\left(f_\phi(\mathbf{x}_z), g_\phi(\mathbf{\hat{y}_i})\right)$. These captions will serve to guide a language model as examples of what the caption to be generated should resemble, through the use of the prompt detailed in Section~\ref{section:llm_prompting}. 
Based on the computed similarities, we retrieve the top-$10$ captions from the datastore. While the final prompt used to query the language model will only contain the top-$k \leq 10$ captions, as determined through experiments, we initially retrieve the top-10 most similar captions considering this pool for the graph-based re-ranking algorithm based on PageRank.

\subsection{Retrieving Few-Shot Examples}
\label{section:few_shot_examples}
To obtain few-shot examples, we rely on the image encoder $f_\phi$ from RS-SigLIP2 to embed the input image $\mathbf{x}_z$, producing the feature vector $f_\phi(\mathbf{x}_z) = \mathbf{v}_z$. Cosine similarities are then computed between $\mathbf{v}_z$ and all image embeddings stored in the datastore $\mathcal{D}$, denoted $\mathbf{\hat{v}}_p$, from which the ten most similar images ${\mathbf{\hat{x}}_p}$ are retrieved. The captions associated to these images are retained to build candidate few-shot demonstrations for the prompt. Ten candidates are selected rather than directly retrieving the final $N$ few-shot examples, so that they can later be reranked through the PageRank procedure described in Section~\ref{section:rerank_pagerank}. 

For each of the retrieved images, we also retrieve the most similar captions from $\mathcal{D}$, using the same cross-modal similarity computation described in Section~\ref{section:ret_similar_capt}, i.e., based on sim$(\mathbf{\hat{x}}_p, \mathbf{\hat{y}}_i) = \cos\left(f_\phi(\mathbf{\hat{x}}_p), g_\phi(\mathbf{\hat{y}}_i)\right)$. The set of ground-truth captions ${\mathbf{\hat{y}}_k}$ associated with each similar image $\mathbf{\hat{x}}_p$ is also ranked according to similarity towards $\mathbf{\hat{x}}_p$. We filter the results so that the ground-truth captions, associated to the examples, differ from the retrieved captions referred to in the previous paragraph. The example images, together with the ranked and filtered similar captions and ground-truth captions, are included in the pool of candidate few-shot examples, and treated as nodes in a semantic-visual graph, which is later processed to refine the prompt composition.

Note that the goal of the few-shot examples is to illustrate how the captioning task should be performed. Consequently, in a multilingual setting, the selected ground-truth captions should be presented in the target language. Since all captions in the datastore $\mathcal{D}$ are originally in English, we employ a machine translation model to convert the selected captions into the target language. Further details regarding the translation approach are provided in Section~\ref{sec:datasets}.

\subsection{Re-ranking with PageRank}
\label{section:rerank_pagerank}
A key novelty of our method is a graph-based re-ranking mechanism designed to enhance the coherence of the retrieved content supplied to the language model. Since our approach relies on retrieved captions and few-shot examples, the quality of this context is paramount. A naive selection of these items can introduce redundancy or inconsistencies. To address this, we adapt the personalized PageRank algorithm~\cite{Page1998PageRank,Gleich2015PageRank}. This allows us to re-rank the retrieved content by considering not only each item's individual relevance to the query image, but also the semantic interrelations within the entire set. We construct a fully connected graph where each retrieved element (e.g., a candidate caption or a few-shot example component) is a node. The input query image, while not a node itself, guides the ranking process through a personalization vector.

The graph's structure is formally defined by a row-stochastic adjacency matrix $W \in \mathbb{R}^{n \times n}$. The weight $W_{ij}$ between two nodes is derived from the cosine similarity of their multimodal embeddings, $\mathbf{e}_i$ and $\mathbf{e}_j$, as produced by RS-SigLIP2:
\begin{equation}
W_{ij} = \frac{\max(\cos(\mathbf{e}_i, \mathbf{e}_j), 0)}{\sum_{k=1}^{n} \max(\cos(\mathbf{e}_i, \mathbf{e}_k), 0)}.
\end{equation}
This ensures that the edge weights are normalized for the stochastic random walk model. The ``personalization'' is achieved through a vector $\mathbf{v}$, which biases the ranking towards nodes most relevant to the input image. Each component $v_i$ is proportional to the similarity $s_i$ between the content of node $i$ and the input image:
\begin{equation}
v_i = \frac{\max(s_i, 0)}{\sum_{j=1}^{n} \max(s_j, 0)}.
\end{equation}
The final scores are captured in the PageRank vector $\mathbf{r}$, calculated as the fixed point of the following iterative update:
\begin{equation}
\mathbf{r} = \alpha W^\top \mathbf{r} + (1 - \alpha) \mathbf{v},
\label{eq:pagerank}
\end{equation}
where $\alpha \in (0,1)$ is a damping factor that balances the influence of the graph's link structure ($W^\top \mathbf{r}$) and the initial relevance encoded in the personalization vector $\mathbf{v}$~\cite{Gleich2015PageRank}.

\begin{figure*}[!htb]
  \centering
  \includegraphics[width=\textwidth]{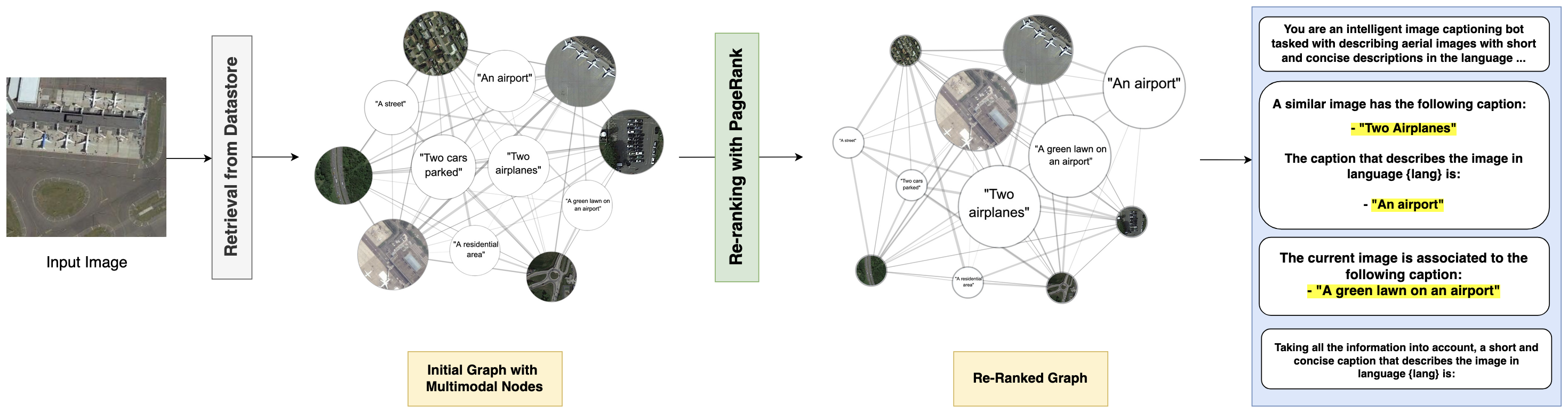}
  \caption[Overview of the PageRank re-ranking process]{An illustration for how the PageRank algorithm influences the final composition of the prompt. On the left, the weighting of the elements is based solely on SigLIP similarities (i.e., the original ranking). On the right, after PageRank propagation, the elements are re-ranked into a different distribution of node weights. For simplicity, this representation considers fewer elements and final retrieved captions compared to the experiments reported in the paper.}
  \label{fig:pagerank_scheme}
\end{figure*}

The resulting score for each node reflects its overall importance, factoring in both its direct relevance to the input image and its semantic connections to other high-quality items. This process effectively promotes a coherent and representative subset of information. Finally, we use the PageRank scores to select the top-$k$ captions and top-$N$ few-shot examples to construct the final prompt, ensuring it is both concise and contextually rich, thereby enhancing the quality of the generated captions. An illustration of the application of PageRank, and the change of node values from SigLIP similarity to final PageRank scores, is given in Figure \ref{fig:pagerank_scheme}.

\subsection{Prompting for Caption Generation}
\label{section:llm_prompting}
\begin{figure}[!ht]
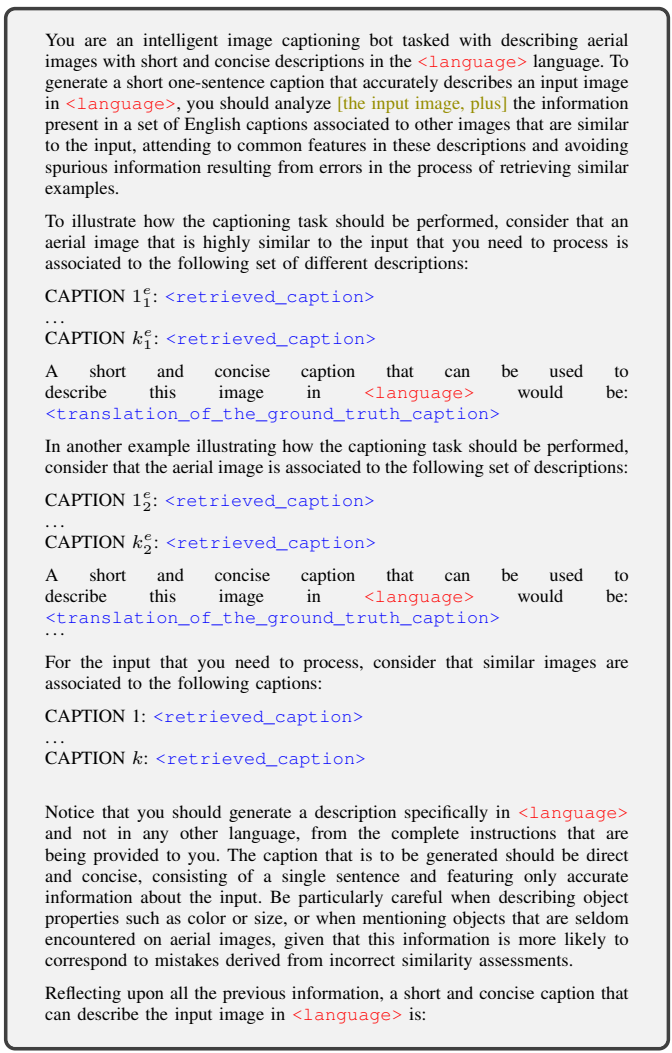

    \centering
\begin{tcolorbox}[enhanced jigsaw, breakable, pad at break*=1mm,fontupper=\scriptsize,fontlower=\scriptsize]
  You are an intelligent image captioning bot tasked with describing aerial images with short and concise descriptions in the {\color{red} \tt <language>} language. To generate a short one-sentence caption that accurately describes an input image in {\color{red} \tt <language>}, you should analyze {\color{olive} [the} {\color{olive} input image, plus]} the information present in a set of English captions associated to other images that are similar to the input, attending to common features in these descriptions and avoiding spurious information resulting from errors in the process of retrieving similar examples.

  \vspace{-0.125cm}
  ~\\
  To illustrate how the captioning task should be performed, consider that an aerial image that is highly similar to the input that you need to process is associated to the following set of different descriptions:

  \vspace{-0.125cm}
  ~\\
  CAPTION $1^e_1$: {\color{blue} \tt <retrieved\_caption>}

  \ldots

  CAPTION $k^e_1$: {\color{blue} \tt <retrieved\_caption>}

  \vspace{-0.125cm}
  ~\\
  A short and concise caption that can be used to describe this image in {\color{red} \tt <language>} would be: {\tt \color{blue}<translation\_of\_the\_ground\_truth\_caption>}

   \vspace{-0.125cm}
   ~\\
   In another example illustrating how the captioning task should be performed, consider that the aerial image is associated to the following set of descriptions:

   \vspace{-0.125cm}
   ~\\
   CAPTION $1^e_2$: {\color{blue} \tt <retrieved\_caption>}

   \ldots

   CAPTION $k^e_2$: {\color{blue} \tt <retrieved\_caption>}

   \vspace{-0.125cm}
   ~\\
    A short and concise caption that can be used to describe this image in {\color{red} \tt <language>} would be: {\color{blue} \tt <translation\_of\_the\_ground\_truth\_caption>}

    \vspace{-0.125cm}
    \ldots
    \vspace{-0.125cm}
    ~\\

    For the input that you need to process, consider that similar images are associated to the following captions:

    \vspace{-0.125cm}
    ~\\
    CAPTION 1: {\color{blue} \tt <retrieved\_caption>}

    \ldots

    CAPTION $k$: {\color{blue} \tt <retrieved\_caption>}

    \vspace{-0.125cm}
    ~\\
    
    Notice that you should generate a description specifically in {\color{red} \tt <language>} and not in any other language, from the complete instructions that are being provided to you. The caption that is to be generated should be direct and concise, consisting of a single sentence and featuring only accurate information about the input. Be particularly careful when describing object properties such as color or size, or when mentioning objects that are seldom encountered on aerial images, given that this information is more likely to correspond to mistakes derived from incorrect similarity assessments.

    \vspace{-0.125cm}
    ~\\
    Reflecting upon all the previous information, a short and concise caption that can describe the input image in {\color{red} \tt <language>} is:
\end{tcolorbox}
\caption[Prompt given to the LLM for image captioning from retrieved captions.]{Prompt for image captioning from the retrieved results. The green text between brackets corresponds to the changes made when using the image-aware strategy, in order to take into account the input image.}
\label{fig:llmprompt}
\end{figure}

The retrieved and re-ranked information is formatted into a textual prompt for a language model with multilingual capabilities. Compared to the prompt used in the original LMCap approach~\cite{ramos2023lmcap}, we adopt a more explicit formulation of the captioning task objective, structured in four main sections. Depending on the strategy being adopted, the prompt is either fed to a text-only multilingual LLM, or to a multimodal VLM. The same overall structure is used across both strategies, allowing for a more controlled comparison between the two. The differences between them are minimal and localized, as illustrated in Figure \ref{fig:llmprompt}.

First, we include a paragraph containing clear instructions regarding the task that we aim the model to perform, and describing the input that will be provided in the following sections. Second, the prompt includes a set of few-shot examples composed by retrieved captions for images similar to the input, illustrating how the captioning task should be performed. Third, we list the retrieved captions for the input image. Finally, the fourth section contains some additional instructions to further guide the model in processing all the information previously presented, and asking the model to generate the final caption. The model uses all this information to generate the final caption in the target language in a few-shot setting, meaning that we do not require any training (i.e., the captions are generated by providing information in the prompt, at inference time, to the language model). 

To ensure that the final prompt remains free of duplicated captions, we apply a filtering procedure using the PageRank-based re-ranked elements. For each input image, the algorithm explores combinations of $k$ retrieved captions drawn from its top-10 candidates, according to the PageRank score hierarchy, giving priority to those with higher scores. These selected captions are marked as used, and for each of the top-10 re-ranked most similar images, the algorithm selects the first unused gold caption as a candidate. It then verifies whether a complete set of $N$ few-shot examples can be formed without redundancy. For each gold candidate, the few-shot retrieved captions are filtered against a dynamic mask containing all previously and potentially used captions. If at least $k$ unique captions remain, the top-$k$ are kept. The procedure stops as soon as a full, repetition-free prompt is successfully built, ensuring that all included captions and examples correspond to the highest-ranking PageRank scores.

\section{Experimental Setup}
\label{sec:exp_setup}

This section summarizes the experimental configuration used in all evaluations. Unless otherwise noted, we set $N=k=3$ (i.e., three few-shot examples and three retrieved captions per example and for the input image). 

For graph-based re-ranking, we use the implementation of personalized PageRank from \texttt{NetworkX}~\cite{hagberg2008networkx} with damping $\alpha=0.9$, i.e. at each iteration there is a 90\% chance of following edges in the multimodal graph (propagating semantic relevance), and a 10\% chance of teleporting to nodes according to the personalization vector, which biases the process toward considering the similarity to the input image.

In addition to our main experimental configuration (with PageRank and the Large Language Models described below), we also report on a set of additional experiments: (i) an experiment with a generalist visual encoder, not fine-tuned to the remote sensing domain; (ii) an ablation without PageRank re-ranking; and (iii) an experiment without the use of retrieved captions and few-shot examples, for the VLM setup.
Image captioning performance was evaluated with standard metrics from the COCOeval package\footnote{\url{https://github.com/salaniz/pycocoevalcap}} with SacreBLEU tokenization\footnote{\url{https://github.com/mjpost/sacrebleu}}, namely through the BLEU1, BLEU4, and CIDEr evaluation metrics, and also with a RefCLIPScore variant~\cite{clipscore}, that uses the RS-SigLIP2 encoder, which we called RefSigLIPScore, and that assesses similarities between the generated caption and (a) the input image plus (b) the ground-truth English caption. The following subsections detail the visual encoder, language models, and datasets that were used.

\subsection{SigLIP2 Visual Encoder and Domain-Specific Fine-Tuning}
We adopt SigLIP2~\cite{tschannen2025siglip2} as visual encoder and retrieval model, using a model derived from the \emph{siglip2-large-patch16-256}\footnote{\url{https://huggingface.co/google/siglip2-large-patch16-256}} version. 
We choose the large variant for its higher representational capacity and stronger alignment performance.

We specifically used a SigLIP2 model that was fine-tuned to the remote sensing domain (RS-SigLIP2) using the same English datasets and training procedure described in connection to RS-M-CLIP~\cite{silva2024multilingual}. The training objective combines (i) a constrastive alignment loss to encourage cross-modal matching, and (ii) self-distillation on the image encoder (based on DINO~\cite{caron2021dino}), where a student matches a teacher updated via Exponential Moving Average (EMA) under diverse augmentations. The final loss averages the alignment and self-distillation terms. 


\subsection{Language Models and Decoding Setup}
In the image-blind variant, the prompt shown in Figure \ref{fig:llmprompt} is fed to a multilingual LLM. We tested the following four different multilingual language models:
\begin{itemize}
  \item \textbf{TowerInstruct-7B-v0.2}\footnote{\url{https://huggingface.co/Unbabel/TowerInstruct-7B-v0.2}}\cite{alves2024tower}: A 7B parameter model that is efficient and fluent for multilingual generation model, instruction-tuned and supporting 10 languages. 
  \item \textbf{EuroLLM-9B/22B-Instruct}\footnote{\url{https://huggingface.co/utter-project/EuroLLM-9B-Instruct}}\textsuperscript{,}\footnote{\url{https://huggingface.co/utter-project/EuroLLM-22B-Instruct-Preview}}~\cite{martins2024eurollm}: Instruction-tuned multilingual models targeting high-quality generation across European and globally relevant languages. We use both the 9B and 22B variants to assess scaling effects.
  \item \textbf{Gemma3-12B-it}\footnote{\url{https://huggingface.co/google/gemma-3-12b-it}}~\cite{gemma3technicalreport}: A recent multilingual instruction-tuned model with strong performance in cross-lingual generation and instruction following.
\end{itemize}
For efficiency, all LLMs are served via HuggingFace Transformers~\cite{wolf2020hftransformers} with 4-bit quantization (bitsandbytes NF4), which substantially reduces memory usage without significantly degrading the generation quality.

For the image-aware variant, we use \textbf{EuroVLM-9B}\footnote{\url{https://huggingface.co/utter-project/EuroVLM-9B-Preview}}, which augments EuroLLM-9B with a visual encoder based on a SigLIP/SigLIP2-style ViT, and \textbf{Gemma3-12B-it} in its VLM configuration. In both cases, visual embeddings are projected into the text latent space and concatenated to the prompt, enabling joint attention over visual and textual context.

Caption generation uses deterministic beam search with a beam width of 3, offering stable and reproducible outputs aligned with the prompt content, which is desirable for automatic metrics (e.g., BLEU, CIDEr) that reward exact $n$-gram matches, while avoiding variance from stochasticity.

\subsection{Remote Sensing Datasets}
\label{sec:datasets}
The remote sensing image captioning datasets that were used in our experiments are described next and also summarized in Table~\ref{table:datasets}.

\begin{itemize}
    \item RSICD features a large set of images collected from different sources~\cite{lu2017exploring}. The images were manually annotated with short captions, but many of the captions are duplicated to ensure 5 captions per image;
	
    \item UCM-Captions and Sydney-Captions are based on two popular remote sensing scene classification datasets that were repurposed for image captioning by manual annotation~\cite{qu2016Deep}. Although extensively used, these datasets are small and have limitations in the fact that the captions correspond to very simple sentences that are highly similar between themselves;
        
	\item NWPU-Captions is currently the largest dataset of images paired to short captions in the remote sensing domain~\cite{cheng2022NWPUCaptions}. Each image is associated to 5 manually annotated captions, and the images span over a large set of classes, describing different land coverage types.
\end{itemize}

\begin{table}[!t]
\centering
\caption{Statistical characterization of the different remote sensing image captioning datasets used in our experiments.}
\label{table:datasets}
\resizebox{\columnwidth}{!}{
\begin{tabular}{l | c | r | c | c | r}
\toprule
\textbf{Dataset} & \textbf{\#Classes} & \textbf{\#Images} & \textbf{Size} & \textbf{Resolution} & \textbf{\#Captions} \\
\midrule
NWPU-Captions~\cite{cheng2022NWPUCaptions} & 45 & 31,500 & $256\times256$ & approx. 30–0.2m & 157,500 \\
RSICD~\cite{lu2017exploring} & 30 & 10,921 & $224\times224$ & var. resolutions & 54,605 \\
Sydney-Captions~\cite{qu2016Deep} & 7 & 613 & $500\times500$ & 0.5m & 3,065 \\
UCM-Captions~\cite{qu2016Deep} & 21 & 2,100 & $256\times256$ & approx. 0.3m & 10,500 \\
\bottomrule
\end{tabular}
}
\end{table}

The training split of the aforementioned datasets was used to build the common datastore from which we retrieve the similar captions and the few-shot examples.

To evaluate captioning performance in different languages, and given the current lack of multilingual datasets for the task of remote sensing image captioning, we translated the captions from the test set instances of all four aforementioned datasets. We used a larger TowerInstruct-13B\footnote{\url{https://huggingface.co/Unbabel/TowerInstruct-13B-v0.1}} language model~\cite{alves2024tower}, whose fine-tuning strategy had a specific focus on machine translation tasks, and which has state-of-the-art performance in several high-resource languages. Given that captions for remote sensing images are often relatively simple sentences, automatic machine translation should not introduce significant errors, thus corresponding to a simple and effective approach that allows us to test multilingual image captioning methods in the remote sensing domain.

The TowerInstruct-13B model was prompted in a zero-shot manner to translate the individual English captions into the nine other languages that are supported by the model, namely German, French, Spanish, Chinese, Portuguese, Italian, Russian, Korean, and Dutch. The following instruction was used to perform the translation of the captions: ``\texttt{Translate the following text from English into \{language\}.\textbackslash nEnglish: \{caption\}\textbackslash n \{language\}:}''

In the proposed method, the few-shot examples are collected with basis on image retrieval, but consist of retrieved English captions together with a ground-truth caption in the target language. The same machine translation approach was used to obtain these translated ground-truth captions. 

Note also that, instead of directly producing captions in a target language, we could also consider generating captions in English, afterwards using the same TowerInstruct model to perform the translation to the target language. However, by directly generating captions in the target language, we avoid this second translation step at inference time, and we can better leverage the capabilities of language models that support a more extensive set of languages. We nonetheless report on an experiment in which we use this two-step generate-then-translate approach, showing that our method performs better.

\section{Results and Discussion}
\label{sec:results}

This section presents the experimental results obtained with the proposed approach and discusses the contribution of each component to the overall performance. We begin by reporting the main results, where our method is evaluated against existing state-of-the-art models, using the complete pipeline described in the previous sections (Section \ref{sec:mainresults}). Next, we conduct several additional experiments to assess the robustness and design choices of the approach. First, we analyze the effect of using a generic visual encoder instead of the fine-tuned one, to isolate the impact of visual adaptation (Section \ref{sec:results_finetuning}). Then, we perform an ablation study where the PageRank-based re-ranking step is removed, allowing us to quantify the gains introduced by this algorithm (Section \ref{sec:impact_pagerank}). We also study the impact of varying the number of retrieved captions ($k$) and few-shot examples ($N$) (Section \ref{sec:ablation_k_N}). Finally, we investigate the influence of the retrieved content on both the prompt composition and the caption generation process, providing a deeper understanding of how external information contributes to model performance (Section \ref{sec:no_retrieved_info}).

\subsection{Main Evaluation Results}
\label{sec:mainresults}

We first report results obtained with the proposed approach across the aforementioned four remote sensing datasets, comparing the image-blind (LLMs) and image-aware (VLMs) variants. Performance is measured with BLEU-1, BLEU-4, and CIDEr, considering both English captions and the average results across nine other languages (Portuguese, Spanish, French, German, Dutch, Italian, Chinese, Korean, and Russian).

Table~\ref{tab:our_models_results} presents the complete results, comparing our models against state-of-the-art approaches from the literature. The results show that TowerInstruct-7B and Gemma3-12B-LM lead to very competitive training-free models, consistently achieving high scores across datasets and languages. EuroLLM-9B also performs strongly, while its larger 22B counterpart underperforms in most cases, showing that parameter scaling does not guarantee better performance in this setup. As expected, multilingual averages are lower than English results, reflecting variability across target languages. Surprisingly, vision–language models achieve slightly lower scores than the text-only LLM counterparts, which suggests that these models generate captions that do not exactly match the style that is used in the considered remote sensing datasets, and that current lexical-overlap metrics may fail to capture the benefits of including image inputs. Importantly, when compared with existing domain-specific large vision and language models, or fully supervised encoder-decoder models, our training-free models remain competitive, matching or surpassing some systems in BLEU and CIDEr, particularly on the UCM, Sydney, and NWPU datasets.

\begin{table*}
\centering
\caption{Performance for both the image-blind (LLMs) and image-aware (VLMs) versions, across the four datasets. In the second part of the table, we provide English captioning results, as reported in prior work involving supervised learning. The best English results are in bold, and the best multilingual results are underlined.}
\label{tab:our_models_results}
\resizebox{\textwidth}{!}{
\begin{tabular}{lcccccccccccc}
\toprule
\multirow{2}{*}{} & \multicolumn{3}{c}{RSICD} & \multicolumn{3}{c}{UCM} & \multicolumn{3}{c}{Sydney} & \multicolumn{3}{c}{NWPU} \\
\cmidrule(lr){2-4} \cmidrule(lr){5-7} \cmidrule(lr){8-10} \cmidrule(lr){11-13}
Model & BLEU1 & BLEU4 & CIDEr & BLEU1 & BLEU4 & CIDEr & BLEU1 & BLEU4 & CIDEr & BLEU1 & BLEU4 & CIDEr \\
\midrule
TowerInstruct-7B (EN)   & 0.607 & 0.249 & 0.567 & \textbf{0.882} & \textbf{0.723} & \textbf{3.462} & 0.778 & 0.574 & \textbf{2.392} & 0.842 & \textbf{0.582} & \textbf{1.573}\\
TowerInstruct-7B (AVG)  & 0.556 & 0.207 & 0.542 & 0.812 & \underline{0.598} & \underline{2.389} & 0.712 & \underline{0.487} & \underline{1.642} & 0.773 & \underline{0.503} & \underline{1.223} \\
\midrule
EuroLLM-9B (EN)      & \textbf{0.646} & \textbf{0.283} & 0.678 & 0.875 & 0.699 & 3.374 & \textbf{0.805} & \textbf{0.579} & 2.315 & 0.855 & 0.561 & 1.503 \\
EuroLLM-9B (AVG)     & 0.558 & 0.212 & 0.560 & 0.813 & 0.591 & 2.327 & 0.722 & 0.483 & 1.504 & 0.762 & 0.469 & 1.091 \\
\midrule
Gemma3-12B-LM (EN)      & 0.632 & 0.256 & 0.691 & 0.867 & 0.681 & 3.170 & 0.774 & 0.521 & 2.050 & \textbf{0.866} & 0.516 & 1.403 \\
Gemma3-12B-LM (AVG)     & \underline{0.574} & \underline{0.214}& \underline{0.576} & \underline{0.821} & 0.549 & 2.131 & 0.738 & 0.460 & 1.426 & \underline{0.796} & 0.427 & 1.011 \\
\midrule
EuroLLM-22B (EN)     & 0.645 & 0.259 & \textbf{0.780} & 0.840 & 0.619 & 3.080 & 0.763 & 0.494 & 1.860 & 0.804 & 0.473 & 1.247 \\
EuroLLM-22B (AVG)    & 0.565 & 0.199 & 0.536 & 0.804 & 0.546 & 2.206 & 0.714 & 0.457 & 1.399 & 0.748 & 0.401 & 0.939 \\
\midrule
EuroVLM-9B (EN)      & 0.571   & 0.213   & 0.605    & 0.870    & 0.687    & 3.435    & 0.768    & 0.555    & 2.295    & 0.788    & 0.479 &  1.321\\
EuroVLM-9B (AVG)     &  0.547  & 0.190   & 0.525    & 0.812    & 0.578    & 2.369    & 0.705    & 0.462    & 1.540    & 0.748    & 0.433 & 1.033 \\
\midrule
Gemma3-12B-VLM (EN)      &  0.613 & 0.245 & 0.578 & 0.879 & 0.697 & 3.221 & 0.769 & 0.531 & 2.066 & 0.851 & 0.505 & 1.351 \\
Gemma3-12B-VLM (AVG)     & 0.562 & 0.194 & 0.531 & 0.811 & 0.535 & 2.034 & \underline{0.741} & 0.462 & 1.383 & 0.774 & 0.384 & 0.925  \\

\midrule\midrule
MLCA-NET~\cite{cheng2022NWPUCaptions}               & 0.757 & 0.461 & 2.356 & 0.826 & 0.668 & 3.240 & 0.831 & 0.580 & 2.324 & 0.745 & 0.478 & 1.164 \\
HCNet~\cite{yang2024hcnet}                  & --    & --    & --    & 0.883 & 0.745 & 3.518 & 0.769 & 0.610 & 2.471 & 0.896 & 0.717 & 2.093 \\
MC-Net~\cite{huang2023mc}                 & 0.728 & 0.433 & 2.454 & 0.845 & 0.679 & 3.355 & 0.834 & 0.607 & 2.564 & 0.741 & 0.478 & 1.159 \\
BITA~\cite{yang2024bootstrapping}                   & 0.774 & 0.504 & \textbf{3.054} & 0.889 & 0.719 & \textbf{3.845} & --    & --    & --    & 0.885    & 0.676    & 1.970 \\
Aware-Transform~\cite{cao2023aware}        & --    & --    & --    & 0.901 & 0.781 & 3.779 & 0.854 & 0.651 & 2.832 & \textbf{0.915} & \textbf{0.750} & \textbf{2.147} \\
Def. Transform~\cite{du2023deformable}         & 0.758 & 0.492 & 2.581 & 0.823 & 0.679 & 3.463 & 0.837 & 0.666 & \textbf{3.037} & 0.752 & 0.483 & 1.207 \\
RSGPT~\cite{hu2023rsgpt}                  & 0.703 & 0.368 & 1.029 & 0.861 & 0.657 & 3.332 & 0.823 & 0.622 & 2.731 & --    & --    & -- \\
SkyEyeGPT~\cite{zhan2024skyeyegpt}             & \textbf{0.867} & \textbf{0.600} & 0.837 & \textbf{0.907} & \textbf{0.784} & 2.368 & \textbf{0.919} & \textbf{0.774} & 1.811 & --    & --    & -- \\
\bottomrule
\end{tabular}
}  
\end{table*}

Since $n$-gram based evaluation metrics focus on surface overlap, they may undervalue captions that are lexically diverse or paraphrastic, but semantically correct. To address this, we additionally report RefSigLIPScores, shown in Table~\ref{tab:siglip_results}. As opposed to BLEU and CIDEr metrics, RefSigLIPScore has not been used before in benchmarks, but it is still useful to understand the alignment between captions and input images. Results indicate that semantic grounding remains stable across models, with the larger LLM, i.e. EuroLLM-22B, often achieving scores comparable to or higher than smaller models, even in cases where $n$-gram metrics decline. This reveals that the choice of evaluation metric plays a crucial role, and we can either analyse the outputs by comparing them with reference captions in terms of how many $n$-grams are matched, or by assessing whether the generated caption is semantically aligned with the image itself.

\begin{table}[!t]
\centering
\caption[Mean RefSigLIPScore values across datasets and models.]{Mean RefSigLIPScore values for each model and dataset, reported for English (EN) and for the average across 9 languages (AVG). The best English results are in bold, and the best multilingual results are underlined.}
\label{tab:siglip_results}
\resizebox{\linewidth}{!}{
\begin{tabular}{lcccc}
\toprule
\multirow{2}{*}{}& \multicolumn{4}{c}{RefSigLIPScore} \\
\cmidrule(l){2-5}
Model & RSICD & UCM & Sydney & NWPU \\
\midrule
TowerInstruct-7B (EN)   & 0.209 & \textbf{0.362} & 0.359 & \textbf{0.368} \\
TowerInstruct-7B (AVG)  & 0.251 & 0.377 & 0.376 & 0.379 \\
\midrule
EuroLLM-9B (EN)       & 0.231 & 0.361 & 0.365 & 0.367 \\
EuroLLM-9B (AVG)      & 0.266 & \underline{0.378} & 0.377 & 0.381 \\
\midrule
Gemma-12B-LM (EN)    & 0.228 & 0.359 & 0.349 & 0.338 \\
Gemma-12B-LM (AVG)   & \underline{0.274} & 0.375 & 0.374 & 0.380 \\
\midrule
EuroLLM-22B (EN)      & \textbf{0.258} & 0.351 & \textbf{0.367} & 0.365 \\
EuroLLM-22B (AVG)     & 0.267 & 0.377 & \underline{0.379} & \underline{0.383} \\
\midrule
EuroVLM-9B (EN)       & 0.225 & 0.360 & 0.360 & 0.351 \\
EuroVLM-9B (AVG)      & 0.254 & 0.373 & 0.373 & 0.372 \\
\midrule
Gemma-12B-VLM (EN)   & 0.220 & 0.355 & 0.356 & 0.337 \\
Gemma-12B-VLM (AVG)  & 0.267 & 0.366 & 0.370 & 0.368 \\
\bottomrule
\end{tabular}
}
\end{table}

\begin{figure*}[!t]
  \centering
  \includegraphics[width=\linewidth]{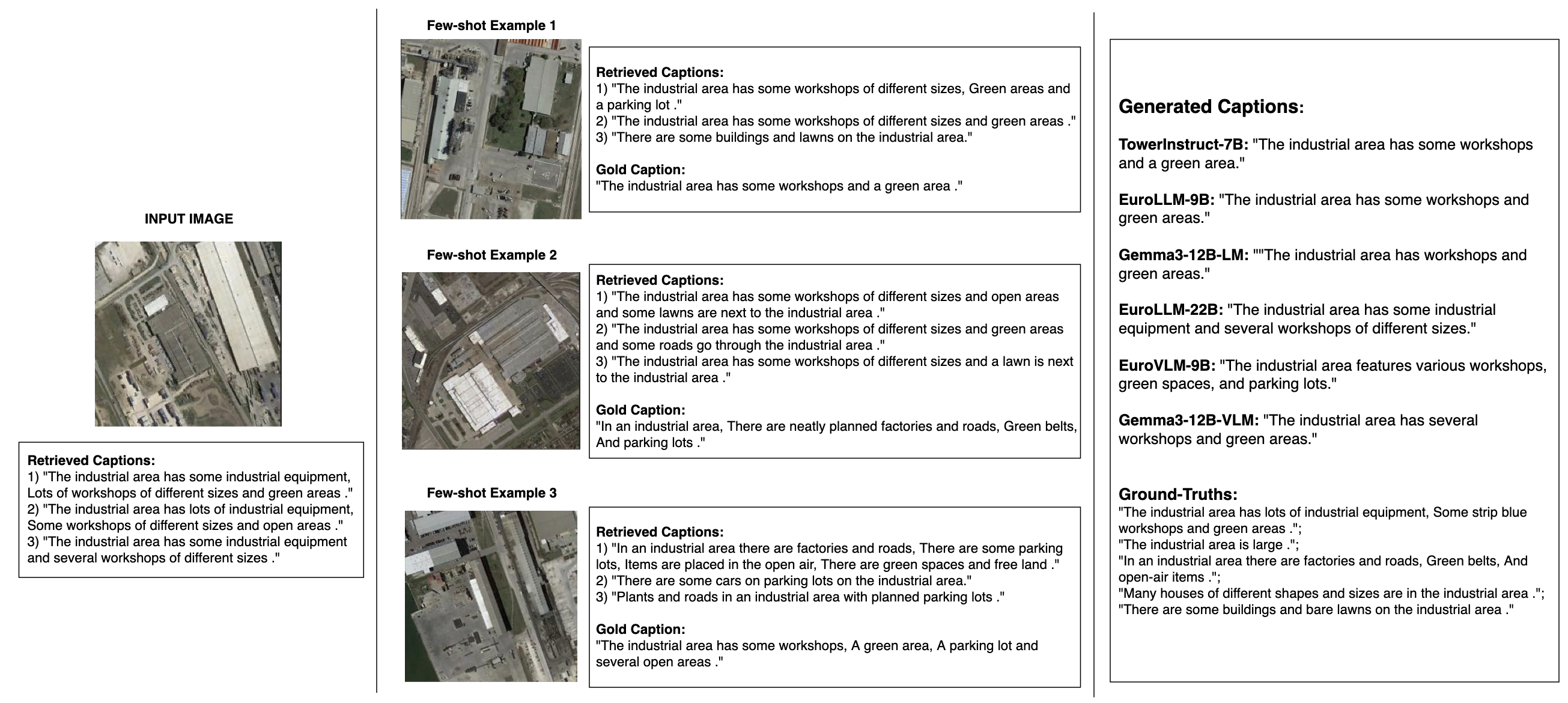}
  \caption[Examples of generated captions from our models, for the English language.]{Examples of generated captions from our models, for the English language and using the input image, the retrieved captions for the input image, and the few-shot examples, with re-ranking through personalized PageRank.}
  \label{fig:final_prompt_example}
\end{figure*}

In order to illustrate how the proposed method worked, and the outputs produced by different models, an example for the results is given in Figure \ref{fig:final_prompt_example}. One can observe the importance of the captions chosen for the prompt, which the models end up using as basis for their generation. This includes not only the retrieved captions for the input image, but also the few-shot examples as well. 

One of the key findings from our results is the performance gap across datasets, with RSICD consistently having the lowest scores across all models. Since this dataset systematically underperforms under our strategy, it is crucial to identify what distinguishes it from the others. To this end, we analyzed the overlap between the $n$-grams appearing in the prompt captions and those in the ground-truth references. A lower overlap implies that the prompt is less informative, providing weaker guidance for the model.

For each input image, the sets of unique 1-grams and 4-grams were extracted from the prompt captions and compared with those from the five reference captions. Overlap was quantified using precision and recall, defined as $P(A,B) = \frac{A \cap B}{A}$ and $R(A,B) = \frac{A \cap B}{B}$, where $A$ denotes the set of prompt $n$-grams and $B$ the set of reference $n$-grams. Mean results for each dataset are given in Table \ref{tab:prompt_reference_overlap}, and they show that RSICD exhibits substantially lower precision and recall compared to the other datasets, indicating weaker prompt–reference alignment. Combined with the fact that only 5\% of RSICD images contain five distinct reference captions, this explains the poorer performance of our strategy. The prompts are less aligned and coherent, and consequently the generated captions align less effectively with the references. 


\begin{table}[h]
\centering
\small
\caption{Prompt--reference overlap (in percentage) using precision and recall metrics, for the four datasets. For each part of the prompt, the 1-gram and 4-gram overlap was computed considering every single caption in the prompt.}
\label{tab:prompt_reference_overlap}
\begin{tabular*}{\linewidth}{@{\extracolsep{\fill}} l c c c c @{}}
\toprule
\multirow{2}{*}{} & \multicolumn{2}{c}{Precision} & \multicolumn{2}{c}{Recall} \\
\cmidrule(l){2-3} \cmidrule(l){4-5}
Dataset & 1-gram & 4-gram & 1-gram & 4-gram \\
\midrule
RSICD  & 38\% &  5\% & 60\% & 13\% \\
UCM    & 60\% & 27\% & 89\% & 64\% \\
Sydney & 58\% & 21\% & 85\% & 48\% \\
NWPU   & 60\% & 21\% & 76\% & 45\% \\
\bottomrule
\end{tabular*}
\end{table}

Another relevant aspect of our strategy is the performance gap between VLMs and their text-only LLM counterparts. Overall, LLM versions tend to achieve better results, with Gemma3 being the model where the two versions are most comparable. The key difference is that VLMs have access to the image, allowing them to introduce image-specific content that would not be inferred from the prompt alone. While this can enrich the generated captions, it also reduces the influence of the prompt. Since the prompt content is generally well aligned with the references, weaker reliance on it often leads to the generation of $n$-grams that are misaligned with both the prompt and the references, ultimately degrading performance.

\begin{figure}[!t]
  \centering
  \includegraphics[width=\linewidth]{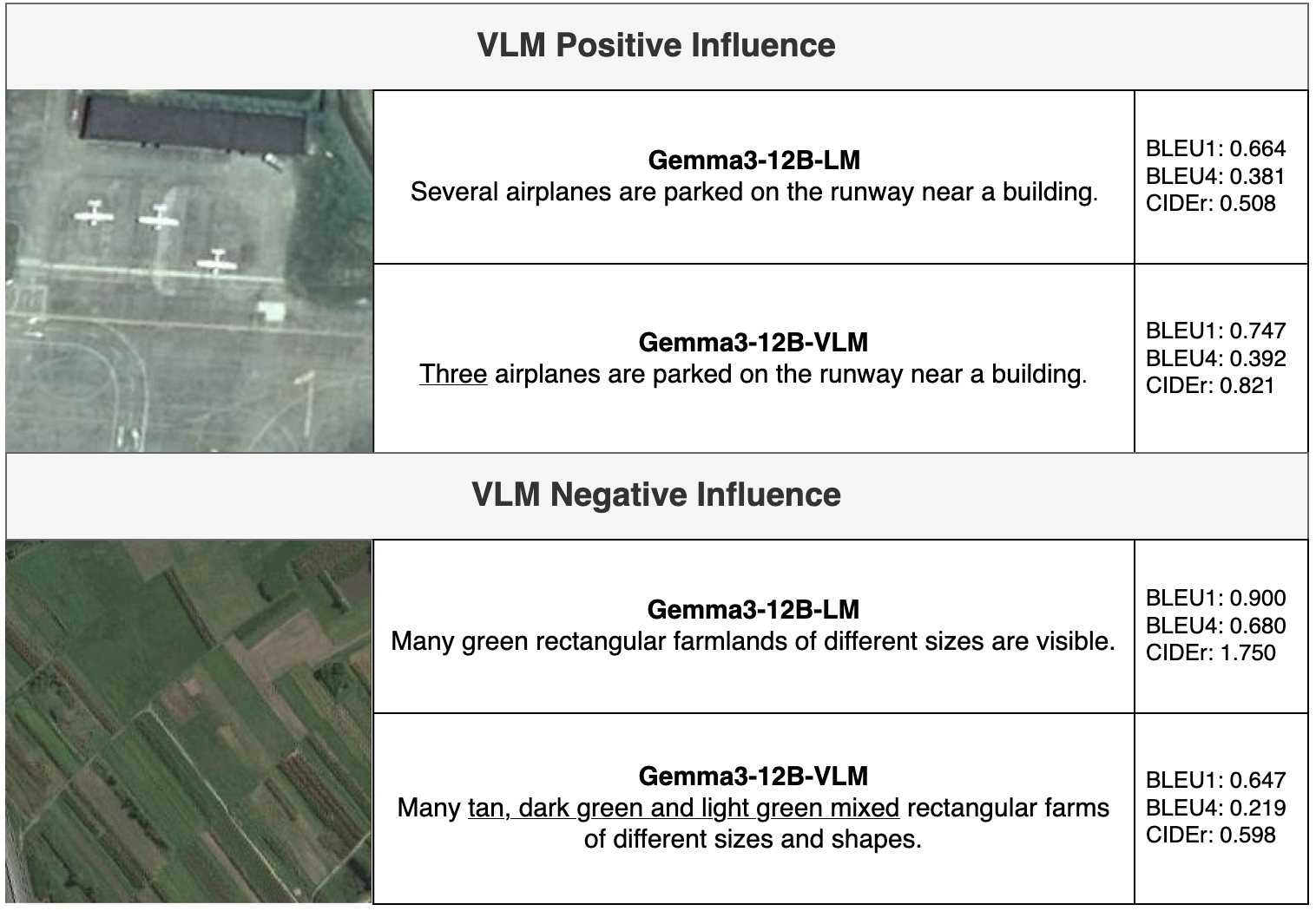}
  \caption[Example for the positive and negative effects of using VLMs.]{Example for the positive and negative effects of using VLMs. On the first case, the input image helps the model clarifying that there are "three" airplanes on the photo, being rewarded by the references. On the other hand, on the second case, the VLM describes in too much detail the image, becoming less aligned with references and having lower scores.}
  \label{fig:example_vlm_lm}
\end{figure}

Table \ref{tab:valid_and_invalid_1grams} reports the distribution of generated 1-grams, occurring or not within the ground-truth references, further segmented by whether they were present in the prompt or not. The results confirm this effect, with VLMs generating more valid 1-grams not included in the prompt, but also producing a substantially higher number of invalid $n$-grams absent from the prompt. This negatively impacts BLEU and CIDEr scores. However, because the generated content still remains visually consistent with the image, the RefSigLIPScore is not significantly affected.
Additionally, these results show that the prompt is the key driver of generation for both VLMs and LLMs, because the majority of the valid (and invalid) 1-grams that are generated are present in the prompt, which translates into a large influence of the prompt on the final generation, independently of having the input image present or not. An example of both the positive or negative influence of using the VLM and the input image, together with the textual prompt, is represented in Figure \ref{fig:example_vlm_lm}.

\begin{table}[!t]
\centering
\caption{Valid and invalid $1$-grams present or absent in the captions of the prompt, for both the LLM and VLM versions of the models that were used.}
\label{tab:valid_and_invalid_1grams}
\resizebox{\linewidth}{!}{
\begin{tabular}{llcccc}
\toprule
\multirow{2}{*}{} & \multirow{2}{*}{} 
& \multicolumn{2}{c}{\textbf{In-prompt}} & \multicolumn{2}{c}{\textbf{Not In-prompt}} \\
\cmidrule(lr){3-4} \cmidrule(lr){5-6}
Dataset & Model & \textit{Valid 1-grams} & \textit{Invalid 1-grams} & \textit{Valid 1-grams} & \textit{Invalid 1-grams} \\
\midrule
\multirow{4}{*}{RSICD}
& EuroLLM-9B     & 7857 & 3290 & 10  & 898  \\
& EuroVLM-9B     & 7927 & 3719 & 101 & 2124 \\
& Gemma3-12B-LM  & 7232 & 2615 & 26  & 776  \\
& Gemma3-12B-VLM & 7078 & 2665 & 43  & 1294 \\
\midrule
\multirow{4}{*}{UCM}
& EuroLLM-9B     & 1977 & 253 & 2 & 14 \\
& EuroVLM-9B     & 2095 & 268 & 0 & 31 \\
& Gemma3-12B-LM  & 1888 & 212 & 1 & 30 \\
& Gemma3-12B-VLM & 1940 & 217 & 3 & 24 \\
\midrule
\multirow{4}{*}{Sydney}
& EuroLLM-9B     & 603 & 142 & 0 & 2 \\
& EuroVLM-9B     & 627 & 174 & 0 & 9 \\
& Gemma3-12B-LM  & 548 & 142 & 0 & 11 \\
& Gemma3-12B-VLM & 560 & 141 & 0 & 18 \\
\midrule
\multirow{4}{*}{NWPU}
& EuroLLM-9B     & 33886 & 5150 & 27 & 174 \\
& EuroVLM-9B     & 34046 & 6186 & 147 & 2257 \\
& Gemma3-12B-LM  & 29414 & 3654 & 57 & 700 \\
& Gemma3-12B-VLM & 29972 & 3944 & 100 & 1059 \\
\bottomrule
\end{tabular}
}
\end{table}

Finally, we present in detail the multilingual results for one of our models, namely Gemma3-12B-LM, in Table \ref{tab:multilingual_results}. Our strategy shows similar performance across languages, with the notable exception of Russian, which performs consistently worse on $n$-gram–based metrics. However, the RefSigLIPScore values are comparable to other languages, suggesting some inconsistencies in evaluation. 

\begin{table*}
\centering
\caption{Multilingual performance of the Gemma3-12B-LM model across datasets. Best results are highlighted in bold.}
\label{tab:multilingual_results}
\resizebox{\textwidth}{!}{
\begin{tabular}{lcccccccccccccccc}
\toprule
 & \multicolumn{4}{c}{RSICD} & \multicolumn{4}{c}{UCM} & \multicolumn{4}{c}{Sydney} & \multicolumn{4}{c}{NWPU} \\
 \cmidrule(lr){2-5} \cmidrule(lr){6-9} \cmidrule(lr){10-13} \cmidrule(lr){14-17}
Language & BLEU1 & BLEU4 & CIDEr & SigLIP & BLEU1 & BLEU4 & CIDEr & SigLIP & BLEU1 & BLEU4 & CIDEr & SigLIP & BLEU1 & BLEU4 & CIDEr & SigLIP \\
\midrule
English    & 0.632 & 0.256 & 0.691 & 0.228 & \textbf{0.867} & \textbf{0.681} & \textbf{3.170} & 0.359 & 0.774 & \textbf{0.521} & \textbf{2.050} & 0.349 & \textbf{0.866} & \textbf{0.516} & \textbf{1.403} & 0.338 \\
Portuguese & 0.582 & 0.223 & 0.628 & 0.266 & 0.823 & 0.567 & 2.392 & 0.377 & 0.751 & 0.496 & 1.577 & 0.376 & 0.797 & 0.438 & 1.131 & 0.380 \\
Spanish    & 0.602 & 0.229 & 0.663 & 0.271 & 0.828 & 0.602 & 2.646 & \textbf{0.383} & 0.744 & 0.509 & 1.507 & 0.377 & 0.833 & 0.483 & 1.231 & 0.381 \\
French     & 0.561 & 0.218 & 0.611 & 0.265 & 0.836 & 0.595 & 2.272 & 0.374 & 0.746 & 0.500 & 1.335 & 0.370 & 0.803 & 0.447 & 1.035 & 0.377 \\
German     & 0.545 & 0.204 & 0.467 & 0.279 & 0.832 & 0.529 & 2.016 & 0.374 & 0.741 & 0.414 & 1.446 & 0.366 & 0.771 & 0.387 & 0.859 & 0.380 \\
Dutch      & 0.591 & 0.207 & 0.564 & 0.269 & 0.832 & 0.528 & 2.072 & 0.372 & 0.785 & 0.493 & 1.504 & 0.368 & 0.820 & 0.427 & 1.039 & 0.382 \\
Italian    & 0.571 & 0.217 & 0.586 & 0.273 & 0.828 & 0.581 & 2.416 & 0.376 & 0.693 & 0.467 & 1.507 & 0.372 & 0.771 & 0.393 & 0.963 & 0.380 \\
Korean     & \textbf{0.670} & \textbf{0.270} & 0.644 & 0.278 & 0.823 & 0.523 & 1.697 & 0.357 & 0.748 & 0.495 & 1.600 & 0.373 & 0.844 & 0.490 & 1.076 & 0.381 \\
Chinese    & 0.600 & 0.238 & \textbf{0.731} & \textbf{0.290} & 0.826 & 0.583 & 2.171 & 0.380 & \textbf{0.783} & 0.480 & 1.509 & \textbf{0.386} & 0.838 & 0.480 & 1.205 & \textbf{0.386} \\
Russian    & 0.448 & 0.118 & 0.292 & 0.273 & 0.758 & 0.431 & 1.496 & 0.382 & 0.653 & 0.285 & 0.847 & 0.378 & 0.688 & 0.301 & 0.562 & 0.377 \\
\bottomrule
\end{tabular}
}
\end{table*}

\begin{figure}[!htb]
  \centering
  \includegraphics[width=\linewidth]{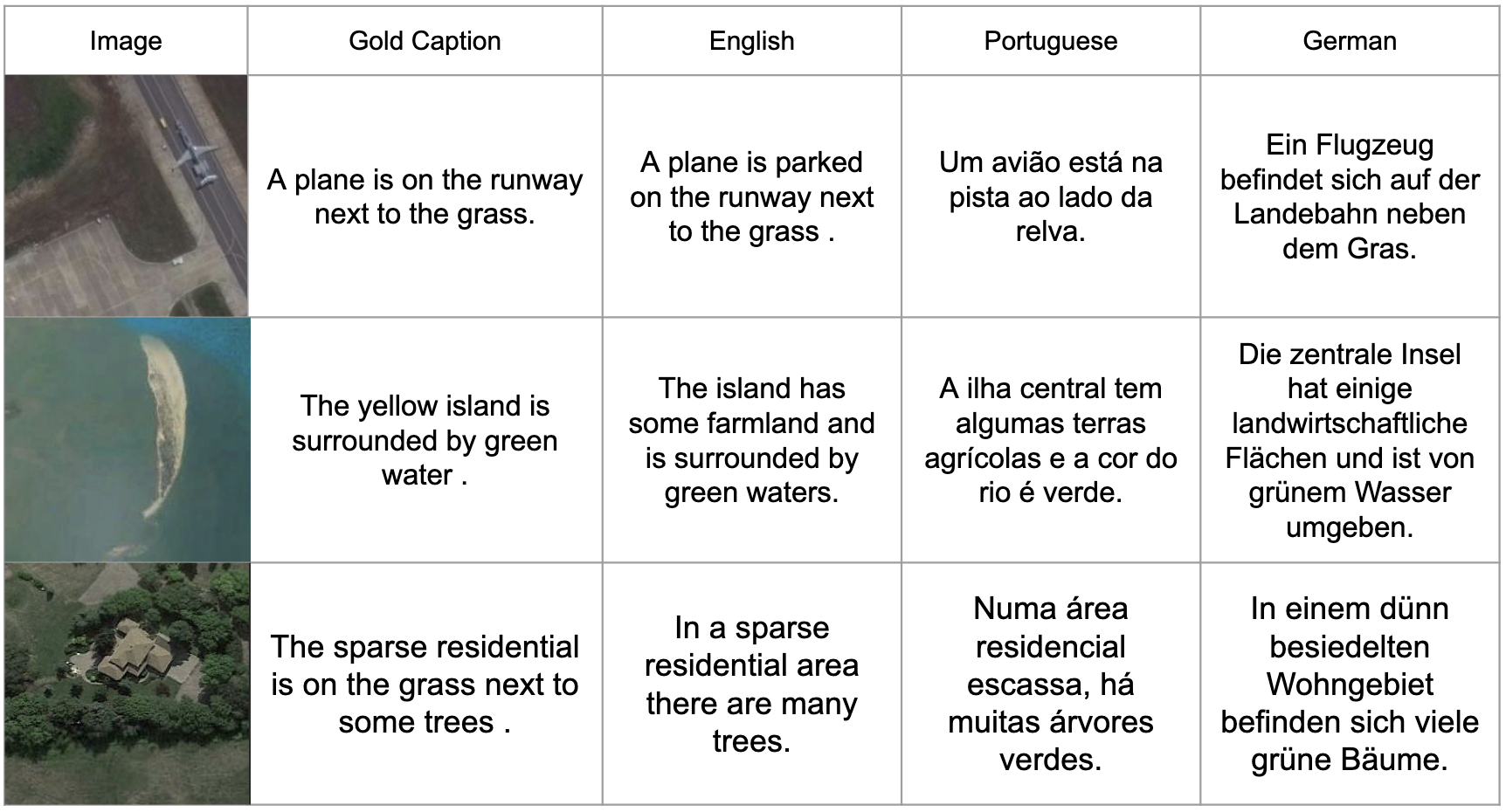}
  \caption[Examples of captions generated with the Gemma3-12B large language model, with the complete method and for three different languages.]{Examples of captions generated with the Gemma3-12B large language model, with the complete method and for three different languages.}
  \label{fig:generative_example}
\end{figure}

An illustrative example of the multilingual performance of Gemma3-12B-LM, with three different target languages and the English ground-truth, is presented in Figure \ref{fig:generative_example}.

One of the reasons for adopting the proposed strategy, instead of generating English captions and translating them into the target language, is that translation introduces an additional computational step and error sources that can degrade caption quality, particularly for $n$-gram–based evaluation metrics. To assess this, we tested a generate-then-translate strategy, applying the same translation procedure to both the generated captions and to the ground-truth. Results are given in Table \ref{tab:translation_vs_multilingual}, showing identical RefSigLIPScore values, as expected since this metric evaluates image–caption alignment and is less sensitive to translation noise. However, BLEU and CIDEr scores are consistently lower under translation, confirming the drawbacks of this approach.

For future work, it would be valuable to build multilingual datasets with human-annotated references, enabling more reliable evaluations and stronger comparisons with alternative image captioning strategies.

\begin{table*}
\centering
\caption{Comparison between multilingual generation and a generate-then-translate approach, across datasets and using the BLEU, CIDEr and RefSigLIPScore metrics. The results are the average of the nine target languages.}
\label{tab:translation_vs_multilingual}
\resizebox{\textwidth}{!}{
\begin{tabular}{lcccccccccccccccc}
\toprule
 & \multicolumn{4}{c}{RSICD} & \multicolumn{4}{c}{UCM} & \multicolumn{4}{c}{Sydney} & \multicolumn{4}{c}{NWPU} \\
 \cmidrule(lr){2-5} \cmidrule(lr){6-9} \cmidrule(lr){10-13} \cmidrule(lr){14-17}
Language & BLEU1 & BLEU4 & CIDEr & SigLIP & BLEU1 & BLEU4 & CIDEr & SigLIP & BLEU1 & BLEU4 & CIDEr & SigLIP & BLEU1 & BLEU4 & CIDEr & SigLIP \\
\midrule
TowerInstruct-7B              & 0.556 & 0.207 & 0.542 & 0.251 & 0.812 & 0.598 & 2.389 & 0.377 & 0.712 & 0.487 & 1.642 & 0.376 & 0.773 & 0.503 & 1.223 & 0.379 \\
TowerInstruct-7B w/ Translation & 0.544 & 0.183 & 0.485 & 0.257 & 0.784 & 0.501 & 1.882 & 0.376 & 0.701 & 0.416 & 1.220 & 0.370 & 0.731 & 0.381 & 0.906 & 0.384 \\
\midrule
EuroLLM-9B                 & 0.558 & 0.212 & 0.560 & 0.266 & 0.813 & 0.591 & 2.327 & 0.378 & 0.722 & 0.483 & 1.504 & 0.377 & 0.762 & 0.469 & 1.091 & 0.381 \\
EuroLLM-9B w/ Translation  & 0.575 & 0.204 & 0.546 & 0.275 & 0.781 & 0.493 & 1.879 & 0.378 & 0.720 & 0.420 & 1.198 & 0.380 & 0.741 & 0.372 & 0.880 & 0.389 \\
\midrule
Gemma-12B-LM               & 0.574 & 0.214 & 0.576 & 0.274 & 0.821 & 0.549 & 2.131 & 0.375 & 0.738 & 0.460 & 1.426 & 0.374 & 0.796 & 0.427 & 1.011 & 0.380 \\
Gemma-12B-LM w/ Translation& 0.565 & 0.194 & 0.531 & 0.267 & 0.791 & 0.493 & 1.800 & 0.378 & 0.697 & 0.393 & 1.068 & 0.371 & 0.755 & 0.353 & 0.859 & 0.375 \\
\midrule
EuroLLM-22B                & 0.565 & 0.199 & 0.536 & 0.267 & 0.804 & 0.546 & 2.206 & 0.377 & 0.714 & 0.457 & 1.399 & 0.379 & 0.748 & 0.401 & 0.939 & 0.383 \\
EuroLLM-22B w/ Translation & 0.547 & 0.185 & 0.505 & 0.274 & 0.743 & 0.436 & 1.715 & 0.372 & 0.687 & 0.388 & 1.007 & 0.379 & 0.696 & 0.318 & 0.736 & 0.383 \\
\midrule
EuroVLM-9B                 & 0.547 & 0.190 & 0.525 & 0.254 & 0.812 & 0.578 & 2.369 & 0.373 & 0.705 & 0.462 & 1.540 & 0.373 & 0.748 & 0.433 & 1.033 & 0.372 \\
EuroVLM-9B w/ Translation  & 0.517 & 0.164 & 0.472 & 0.263 & 0.767 & 0.470 & 1.832 & 0.375 & 0.699 & 0.420 & 1.189 & 0.371 & 0.686 & 0.319 & 0.773 & 0.370 \\
\midrule
Gemma-12B-VLM              & 0.562 & 0.194 & 0.531 & 0.267 & 0.811 & 0.535 & 2.034 & 0.366 & 0.741 & 0.462 & 1.383 & 0.370 & 0.774 & 0.384 & 0.925 & 0.368 \\
Gemma-12B-VLM w/ Translation& 0.563 & 0.193 & 0.520 & 0.271 & 0.786 & 0.494 & 1.800 & 0.377 & 0.695 & 0.399 & 1.095 & 0.372 & 0.743 & 0.343 & 0.831 & 0.373 \\
\bottomrule
\end{tabular}
}
\end{table*}

\subsection{Impact of Fine-tuning the Visual Encoder}
\label{sec:results_finetuning}

Following the experiments presented earlier, we further investigated the impact of using RS-SigLIP2 as the visual encoder, since in our main setup this model was fine-tuned with the domain data. To better understand the contribution of fine-tuning, we designed a comparable setup where retrieval was performed using the original SigLIP2 model, without any domain-specific fine-tuning. The results of this comparison are shown in Table \ref{tab:our_models_results_nft}.

\begin{table*}[!h]
\centering
\caption{Experiments with a SigLIP2 model without fine-tuning for the remote sensing domain, for every language model that was tested on the main experiments.}
\label{tab:our_models_results_nft}
\resizebox{\textwidth}{!}{
\begin{tabular}{lcccccccccccccccc}
\toprule
\multirow{2}{*}{} & \multicolumn{4}{c}{RSICD} & \multicolumn{4}{c}{UCM} & \multicolumn{4}{c}{Sydney} & \multicolumn{4}{c}{NWPU} \\
\cmidrule(lr){2-5} \cmidrule(lr){6-9} \cmidrule(lr){10-13} \cmidrule(lr){14-17}
Model & BLEU1 & BLEU4 & CIDEr & SigLIP & BLEU1 & BLEU4 & CIDEr & SigLIP & BLEU1 & BLEU4 & CIDEr & SigLIP & BLEU1 & BLEU4 & CIDEr & SigLIP \\
\midrule
TowerInstruct-7B (EN)   & 0.552 & 0.191 & 0.453 & 0.132 & 0.773 & 0.568 & 2.728 & 0.282 & 0.690 & 0.408 & 1.555 & 0.289 & 0.746 & 0.421 & 1.171 & 0.255 \\
TowerInstruct-7B (AVG)  & 0.526 & 0.182 & 0.472 & 0.194 & 0.766 & 0.546 & 2.104 & 0.331 & 0.686 & 0.409 & 1.253 & 0.349 & 0.713 & 0.404 & 0.953 & 0.308 \\
\midrule
EuroLLM-9B (EN)         & 0.515 & 0.138 & 0.383 & 0.115 & 0.724 & 0.493 & 2.420 & 0.247 & 0.694 & 0.393 & 1.455 & 0.271 & 0.721 & 0.363 & 1.033 & 0.222 \\
EuroLLM-9B (AVG)        & 0.500 & 0.163 & 0.426 & 0.173 & 0.738 & 0.499 & 1.955 & 0.300 & 0.678 & 0.390 & 1.138 & 0.323 & 0.708 & 0.376 & 0.886 & 0.284 \\
\midrule
Gemma3-12B-LM (EN)      & 0.537 & 0.147 & 0.515 & 0.155 & 0.734 & 0.485 & 2.272 & 0.247 & 0.694 & 0.371 & 1.180 & 0.235 & 0.776 & 0.371 & 1.044 & 0.207 \\
Gemma3-12B-LM (AVG)     & 0.479 & 0.135 & 0.379 & 0.169 & 0.691 & 0.385 & 1.451 & 0.259 & 0.630 & 0.319 & 0.826 & 0.278 & 0.714 & 0.316 & 0.741 & 0.249 \\
\midrule
EuroLLM-22B (EN)        & 0.456 & 0.089 & 0.338 & 0.091 & 0.554 & 0.258 & 1.132 & 0.146 & 0.597 & 0.309 & 1.171 & 0.228 & 0.640 & 0.263 & 0.754 & 0.193 \\
EuroLLM-22B (AVG)       & 0.465 & 0.111 & 0.333 & 0.132 & 0.614 & 0.324 & 1.161 & 0.217 & 0.615 & 0.301 & 0.846 & 0.262 & 0.633 & 0.268 & 0.643 & 0.243 \\
\midrule
EuroVLM-9B (EN)         & 0.441 & 0.108 & 0.354 & 0.138 & 0.667 & 0.426 & 2.142 & 0.241 & 0.644 & 0.387 & 1.710 & 0.314 & 0.671 & 0.319 & 0.934 & 0.240 \\
EuroVLM-9B (AVG)        & 0.488 & 0.137 & 0.426 & 0.192 & 0.726 & 0.483 & 1.945 & 0.311 & 0.671 & 0.379 & 1.182 & 0.351 & 0.677 & 0.334 & 0.813 & 0.295 \\
\midrule
Gemma3-12B-VLM (EN)     & 0.529 & 0.163 & 0.440 & 0.153 & 0.736 & 0.481 & 2.251 & 0.250 & 0.687 & 0.410 & 1.363 & 0.261 & 0.772 & 0.370 & 1.051 & 0.223 \\
Gemma3-12B-VLM (AVG)    & 0.476 & 0.124 & 0.365 & 0.178 & 0.682 & 0.362 & 1.300 & 0.260 & 0.650 & 0.337 & 0.861 & 0.281 & 0.702 & 0.292 & 0.703 & 0.256 \\
\bottomrule
\end{tabular}
}
\end{table*}

Overall, the non–fine-tuned setup exhibits systematically lower performance, which is consistent with expectations. Fine-tuning adapts the visual encoder to the remote sensing domain, allowing it to identify captions that more accurately match each target image in the datastore. This improvement propagates through the different steps in the proposed approach, as higher-quality retrieval leads to stronger prompts, which in turn guide the generation process more effectively. To confirm this effect, we performed a prompt–reference overlap analysis, similar to that described in the previous section. The results, presented in Table \ref{tab:overlap_prompt_ref_nft}, show that the fine-tuned encoder achieves higher precision and recall than the non–fine-tuned encoder. In contrast, the lack of fine-tuning produces prompts that are less aligned with the references, which directly results in weaker caption generation and lower final performance.

\begin{table}[!t]
\centering
\small
\caption{Prompt–reference overlap (\%) for Fine-Tuned (FT) and Non Fine-Tuned (NFT) visual encoders, per dataset.}
\label{tab:overlap_prompt_ref_nft}
\resizebox{\linewidth}{!}{
\begin{tabular}{lccccccccc}
\toprule
\multirow{2}{*}{} & \multicolumn{2}{c}{$1-$gram Precision} & \multicolumn{2}{c}{$4-$gram Precision} & \multicolumn{2}{c}{$1-$gram Recall} & \multicolumn{2}{c}{$4-$gram Recall} \\
 \cmidrule(l){2-3} \cmidrule(l){4-5} \cmidrule(l){6-7} \cmidrule(l){8-9}
Dataset & FT & NFT &  FT & NFT &  FT & NFT &  FT & NFT \\
\midrule
RSICD & 38\% & 27\% & 5\% & 2\% & 60\% & 61\% & 13\% & 8\%   \\
UCM  & 60\% & 30\% & 27\% & 10\% & 89\% & 82\% & 64\% & 41\%  \\
Sydney & 58\% & 33\% & 21\% & 10\% & 85\% & 83\% & 48\% & 36\% \\
NWPU & 60\% & 43\% & 21\% & 10\% & 76\% & 73\% & 45\% & 27\% \\
\bottomrule
\end{tabular}
}
\end{table}

An additional aspect worth analyzing is the effect of this change on VLMs. One might initially expect these models to be more robust to poorer retrieval, since they have direct access to the input image. However, our results indicate otherwise. Despite having visual information available, VLMs still rely heavily on the textual prompt as the primary driver for text generation. The majority of the generated $n$-grams, whether correct or incorrect, seem to originate from the prompt. Consequently, when prompt quality deteriorates due to poorer retrieval, VLMs experience a performance drop similar to that of LLMs. This reinforces the conclusion that, in our strategy, the prompt remains the key element for achieving strong performance, with fine-tuning of the visual encoder playing a decisive role in ensuring that the prompt is both informative and aligned with the references.

\subsection{Ablation Experiments}

\subsubsection{Impact of PageRank Re-ranking}
\label{sec:impact_pagerank}
To assess the impact of PageRank re-ranking on our approach, we designed an alternative setup where this step was omitted. In this configuration, the order of retrieval was determined solely by the similarity scores from RS-SigLIP2, both for the retrieved captions and for the ordering of the few-shot examples. This setup was applied to both the LLM and VLM strategies in order to examine whether the effect differed between the two variants. Results are summarized in Table \ref{tab:ablation_PR}.

\begin{table*}[!ht]
\centering
\small
\caption[Results for the ablation study focused on re-ranking with PageRank, for the the two approaches.]{Results for the ablation study focused on re-ranking with PageRank, for the two alternative approaches.}
\label{tab:ablation_PR}
\resizebox{\textwidth}{!}{
\begin{tabular}{lcc cccc cccc cccc cccc}
\toprule
\multirow{2}{*}{} & \multirow{2}{*}{} & \multirow{2}{*}{} & \multicolumn{4}{c}{RSICD} & \multicolumn{4}{c}{UCM} & \multicolumn{4}{c}{Sydney} & \multicolumn{4}{c}{NWPU} \\
\cmidrule(lr){4-7} \cmidrule(lr){8-11} \cmidrule(lr){12-15} \cmidrule(lr){16-19}
Model & PageRank & Language & BLEU1 & BLEU4 & CIDEr & SigLIP & BLEU1 & BLEU4 & CIDEr & SigLIP & BLEU1 & BLEU4 & CIDEr & SigLIP & BLEU1 & BLEU4 & CIDEr & SigLIP \\
\midrule
\multirow{4}{*}{EuroLLM-9B} & \multirow{2}{*}{No}  & EN  & 0.622 & 0.237 & 0.606 & 0.220 & 0.859 & 0.668 & 3.221 & 0.353 & 0.764 & 0.473 & 1.798 & 0.345 & 0.843 & 0.520 & 1.431 & 0.343 \\ 
                         &                      & AVG & 0.543 & 0.191 & 0.520 & 0.262 & 0.789 & 0.557 & 2.215 & 0.377 & 0.660 & 0.384 & 1.150 & 0.364 & 0.722 & 0.427 & 0.937 & 0.364 \\ 
                         & \multirow{2}{*}{Yes} & EN  & 0.646 & 0.283 & 0.678 & 0.231 & 0.875 & 0.699 & 3.374 & 0.361 & 0.805 & 0.579 & 2.315 & 0.365 & 0.855 & 0.561 & 1.503 & 0.367 \\ 
                         &                      & AVG & 0.558 & 0.212 & 0.560 & 0.266 & 0.813 & 0.591 & 2.327 & 0.378 & 0.722 & 0.483 & 1.504 & 0.377 & 0.762 & 0.469 & 1.091 & 0.381 \\ 
\midrule
\multirow{4}{*}{EuroVLM-9B}  & \multirow{2}{*}{No}  & EN  & 0.518 & 0.167 & 0.518 & 0.218 & 0.841 & 0.642 & 3.218 & 0.348 & 0.734 & 0.469 & 1.849 & 0.347 & 0.747 & 0.409 & 1.159 & 0.323 \\ 
                          &                       & AVG & 0.513 & 0.160 & 0.466 & 0.251 & 0.781 & 0.533 & 2.205 & 0.368 & 0.677 & 0.401 & 1.286 & 0.364 & 0.709 & 0.363 & 0.884 & 0.351 \\ 
                          & \multirow{2}{*}{Yes} & EN  & 0.571 & 0.213 & 0.605 & 0.225 & 0.870 & 0.687 & 3.435 & 0.360 & 0.768 & 0.555 & 2.295 & 0.360 & 0.788 & 0.479 & 1.321 & 0.351 \\ 
                          &                       & AVG & 0.547 & 0.190 & 0.525 & 0.254 & 0.812 & 0.578 & 2.369 & 0.373 & 0.705 & 0.462 & 1.540 & 0.373 & 0.748 & 0.433 & 1.033 & 0.372 \\ 
\midrule
\multirow{4}{*}{GemmaLM} & \multirow{2}{*}{No}  & EN  & 0.595 & 0.220 & 0.632 & 0.226 & 0.848 & 0.656 & 3.066 & 0.351 & 0.725 & 0.439 & 1.519 & 0.338 & 0.850 & 0.472 & 1.289 & 0.313 \\ 
                         &                      & AVG & 0.546 & 0.185 & 0.531 & 0.269 & 0.800 & 0.529 & 2.023 & 0.373 & 0.697 & 0.382 & 1.082 & 0.365 & 0.776 & 0.383 & 0.915 & 0.365 \\ 
                         & \multirow{2}{*}{Yes} & EN  & 0.632 & 0.256 & 0.691 & 0.228 & 0.867 & 0.681 & 3.170 & 0.359 & 0.774 & 0.521 & 2.050 & 0.349 & 0.866 & 0.516 & 1.403 & 0.338 \\ 
                         &                      & AVG & 0.574 & 0.214 & 0.576 & 0.274 & 0.821 & 0.549 & 2.131 & 0.375 & 0.738 & 0.460 & 1.426 & 0.374 & 0.796 & 0.427 & 1.011 & 0.380 \\ 
\midrule
\multirow{4}{*}{GemmaVLM} & \multirow{2}{*}{No}  & EN  & 0.591 & 0.211 & 0.554 & 0.216 & 0.858 & 0.669 & 3.116 & 0.349 & 0.747 & 0.459 & 1.710 & 0.354 & 0.838 & 0.465 & 1.265 & 0.316 \\ 
                          &                        & AVG & 0.535 & 0.168 & 0.487 & 0.265 & 0.791 & 0.506 & 1.955 & 0.364 & 0.702 & 0.389 & 1.142 & 0.360 & 0.750 & 0.337 & 0.818 & 0.355 \\ 
                          & \multirow{2}{*}{Yes}  & EN  & 0.613 & 0.245 & 0.578 & 0.220 & 0.879 & 0.697 & 3.221 & 0.355 & 0.769 & 0.531  & 2.066 & 0.356 & 0.851 & 0.505 & 1.351 & 0.337 \\ 
                          &                        & AVG & 0.562 & 0.194 & 0.531 & 0.267 & 0.811 & 0.535 & 2.034 & 0.366 & 0.741 & 0.462 & 1.383 & 0.370 & 0.774 & 0.384 & 0.925 & 0.368 \\ 
\bottomrule
\end{tabular}
}
\end{table*}

The results reveal that the inclusion of PageRank re-ranking provides a clear improvement, particularly in terms of $n$-gram–based metrics such as BLEU and CIDEr. This effect is analogous to what was observed in the previous section when comparing fine-tuned and non fine-tuned visual encoders. Ultimately, PageRank contributes significantly to the quality of the prompt content provided to the models.

A prompt–reference overlap analysis, shown in Table \ref{tab:overlap_prompt_ref_npr}, further illustrates this behavior. PageRank improves alignment by selecting a more coherent ordering of the retrieved captions, leading to prompts that are more consistent with the references. As a result, the final performance lies between that of the main setup (with fine-tuned retrieval plus PageRank) and the non fine-tuned retrieval baseline. These findings confirm that PageRank re-ranking plays a central role as a fine-tuning mechanism for prompt construction, refining the retrieved content and enhancing overlap with the references. At the same time, its effectiveness relies on the strength of the underlying retrieval step, highlighting that both high-quality visual encoding and re-ranking are necessary to achieve optimal results.

\begin{table}[!t]
\centering
\small
\caption{Prompt–reference overlap (\%) for the PageRank (PR) and no PageRank (NPR) cases, per dataset.}
\label{tab:overlap_prompt_ref_npr}
\resizebox{\linewidth}{!}{
\begin{tabular}{lccccccccc}
\toprule
\multirow{2}{*}{} & \multicolumn{2}{c}{$1-$gram Precision} & \multicolumn{2}{c}{$4-$gram Precision} & \multicolumn{2}{c}{$1-$gram Recall} & \multicolumn{2}{c}{$4-$gram Recall} \\
 \cmidrule(l){2-3} \cmidrule(l){4-5} \cmidrule(l){6-7} \cmidrule(l){8-9}
Dataset & PR & NPR &  PR & NPR &  FT & NPR &  PR & NPR \\
\midrule
RSICD & 38\% & 34\% & 5\% & 4\% & 60\% & 64\% & 13\% & 13\% \\
UCM   & 60\% & 57\% & 27\% & 26\% & 89\% & 90\% & 64\% & 64\% \\
Sydney& 58\% & 55\% & 21\% & 19\% & 85\% & 85\% & 48\% & 45\% \\
NWPU  & 60\% & 56\% & 21\% & 18\% & 76\% & 77\% & 45\% & 40\% \\
\bottomrule
\end{tabular}
}
\end{table}

\begin{figure*}[!htb]
  \centering
  \includegraphics[width=\textwidth]{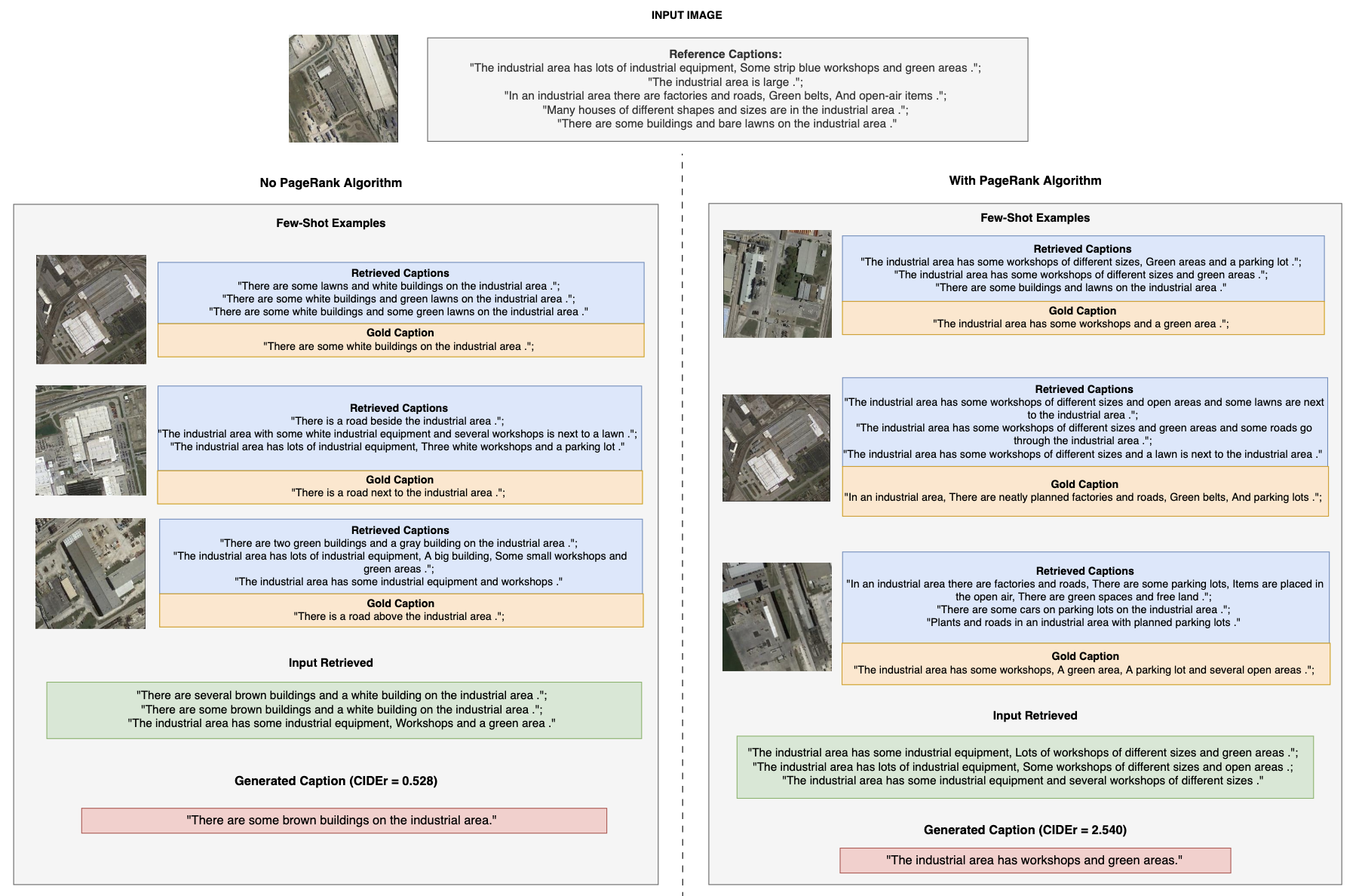}
  \caption[Example of a prompt with and without the use of the reranking PageRank algorithm.]{Examples of prompts constructed with and without the use of the PageRank re-ranking algorithm, where the CIDEr score of the final generated caption has a very different value for the two cases.}

  \label{fig:pr_vs_npr_example}
\end{figure*}

A full example illustrating the impact of PageRank is given in Figure \ref{fig:pr_vs_npr_example}, where we can see that the re-ranking mechanism changes completely the few-shot examples, the retrieved captions, and consequently also the final output.

\subsubsection{Impact of the Number of Retrieved Captions and Few-shot Examples}
\label{sec:ablation_k_N}

In our retrieval-augmented captioning framework, the number of retrieved captions ($k$) and the number of few-shot examples included in the prompt ($N$) jointly determine how much external semantic context is provided to the language model. In the main experiments, we adopted the configuration $N$ = $k$ = 3, which balances prompt length and retrieval diversity. However, increasing the amount of retrieved information may further improve the model’s ability to align image content with relevant caption patterns, at the cost of a longer prompt and potentially higher redundancy.

To assess the sensitivity of the proposed method to these parameters, we extend the analysis to two additional configurations: $N$ = $k$ = 4 and $N$ = $k$ = 5. Evaluating these larger settings allows us to understand whether retrieval diversity and prompt density continue to translate into performance gains, or whether diminishing returns emerge as the prompt grows.

Table \ref{tab:ablation_N_K} reports the results obtained with the three configurations ($N$ = $k$ = 3, 4, 5), across the image-blind and image-aware versions. The results indicate a consistent improvement in captioning performance as $N$ and $k$ increase, although one should also note that this comes with higher costs in terms of LLM processing. The observed trend suggests that enlarging both the retrieved caption set and the few-shot context provides the language model with a richer pool of semantic cues, which in turn enhances the relevance and grounding of the generated captions. To better understand the mechanism behind this improvement, we performed a prompt–reference overlap analysis.

\begin{table*}[!t]
\centering
\small
\caption{Results for the ablation study focused on varying the number of retrieved captions ($k$) and the number of few-shot examples ($N$), for the image-blind and image-aware approaches. For each model, the best English results are highlighted in bold, and the best multilingual results are underlined.}
\label{tab:ablation_N_K}
\resizebox{\textwidth}{!}{
\begin{tabular}{lcc cccc cccc cccc cccc}
\toprule
\multirow{2}{*}{} & \multirow{2}{*}{} & \multirow{2}{*}{} & \multicolumn{4}{c}{RSICD} & \multicolumn{4}{c}{UCM} & \multicolumn{4}{c}{Sydney} & \multicolumn{4}{c}{NWPU} \\
\cmidrule(lr){4-7} \cmidrule(lr){8-11} \cmidrule(lr){12-15} \cmidrule(lr){16-19}
Model & $N$, $k$ & Language & BLEU1 & BLEU4 & CIDEr & SigLIP & BLEU1 & BLEU4 & CIDEr & SigLIP & BLEU1 & BLEU4 & CIDEr & SigLIP & BLEU1 & BLEU4 & CIDEr & SigLIP \\
\midrule
\multirow{6}{*}{EuroLLM-9B}
& \multirow{2}{*}{3} & EN & 0.646 & 0.283 & 0.678 & 0.231 & 0.875 & 0.699 & 3.374 & \textbf{0.361} & \textbf{0.805} & \textbf{0.579} & \textbf{2.315} & 0.365 & 0.855 & 0.561 & 1.503 & \textbf{0.367} \\ 
                         &                      & AVG & 0.558 & 0.212 & 0.560 & \underline{0.266} & \underline{0.813} & 0.591 & \underline{2.327} & \underline{0.378} & 0.722 & 0.483 & \underline{1.504} & 0.377 & 0.762 & 0.469 & 1.091 & \underline{0.381}  \\
  \cmidrule(lr){2-19}
  & \multirow{2}{*}{4} & EN  & \textbf{0.652} & \textbf{0.286} & \textbf{0.697} & \textbf{0.233} & \textbf{0.884} & \textbf{0.706} & \textbf{3.448} & \textbf{0.361} & 0.785 & 0.564 & 2.212 & 0.364 & \textbf{0.859} & 0.572 & 1.537 & 0.366 \\ 
  &                     & AVG & 0.563 & 0.217 & 0.569 & 0.265 & 0.812 & \underline{0.594} & 2.305 & \underline{0.378} & 0.719 & 0.479 & 1.432 & \underline{0.380} & 0.772 & 0.479 & 1.123 & 0.380 \\ 
  \cmidrule(lr){2-19}
  & \multirow{2}{*}{5} & EN  & 0.649 & 0.285 & 0.680 & 0.228 & 0.864 & 0.681 & 3.234 & 0.359 & 0.798 & 0.560 & 2.174 & \textbf{0.367} & \textbf{0.859} & \textbf{0.574} & \textbf{1.546} & 0.364 \\ 
  &                    & AVG & \underline{0.575} & \underline{0.224} & \underline{0.588} & \underline{0.266} & 0.801 & 0.583 & 2.286 & 0.376 & \underline{0.737} & \underline{0.484} & 1.408 & 0.378 & \underline{0.785} & \underline{0.489} & \underline{1.151} & 0.379 \\
\midrule
\multirow{6}{*}{EuroVLM-9B} 
    & \multirow{2}{*}{3} & EN  & 0.571 & 0.213 & 0.605 & 0.225 & 0.870 & 0.687 & 3.435 & \textbf{0.360} & \textbf{0.768} & \textbf{0.555} & \textbf{2.295} & \textbf{0.360} & 0.788 & 0.479 & 1.321 & 0.351 \\ 
                          &                       & AVG & 0.547 & 0.190 & 0.525 & 0.254 & 0.812 & 0.578 & \underline{2.369} & 0.373 & \underline{0.705} & \underline{0.462} & \underline{1.540} & 0.373 & 0.748 & 0.433 & 1.033 & 0.372  \\
  \cmidrule(lr){2-19}
  & \multirow{2}{*}{4} & EN  & 0.596 & 0.227 & 0.675 & 0.237 & \textbf{0.873} & \textbf{0.691} & \textbf{3.446} & \textbf{0.360} & 0.761 & \textbf{0.555} & 2.294 & 0.357 & 0.808 & 0.509 & 1.388 & 0.356 \\ 
  &                     & AVG & 0.552 & 0.195 & 0.532 & 0.253 & \underline{0.813} & \underline{0.579} & 2.353 & \underline{0.374} & 0.700 & 0.457 & 1.532 & 0.373 & 0.755 & 0.448 & 1.069 & \underline{0.373} \\ 
\cmidrule(lr){2-19}
& \multirow{2}{*}{5} & EN  & \textbf{0.621} & \textbf{0.252} & \textbf{0.736} & \textbf{0.242} & 0.862 & 0.670 & 3.282 & 0.357 & 0.754 & 0.547 & 2.278 & 0.359 & \textbf{0.815} & \textbf{0.518} & \textbf{1.403} & \textbf{0.358} \\ 
  &                    & AVG & \underline{0.558} & \underline{0.200} & \underline{0.543} & \underline{0.255} & 0.801 & 0.563 & 2.265 & 0.373 & \underline{0.705} & 0.460 & 1.493 & \underline{0.375} & \underline{0.759} & \underline{0.454} & \underline{1.082} & \underline{0.373} \\

\midrule
\multirow{6}{*}{GemmaLM}
  & \multirow{2}{*}{3} & EN  & 0.632 & 0.256 & 0.691 & 0.228 & 0.867 & 0.681 & 3.170 & \textbf{0.359} & 0.774 & 0.521 & 2.050 & 0.349 & 0.866 & 0.516 & 1.403 & 0.338 \\ 
                         &                      & AVG & 0.574 & 0.214 & 0.576 & 0.274 & 0.821 & 0.549 & \underline{2.131} & 0.375 & 0.738 & 0.460 & \underline{1.426} & 0.374 & 0.796 & 0.427 & 1.011 & 0.380  \\
  \cmidrule(lr){2-19}
  & \multirow{2}{*}{4} & EN  & 0.658 & 0.281 & 0.786 & 0.244 & \textbf{0.879} & \textbf{0.700} & 3.211 & 0.354 & 0.774 & 0.508 & 1.936 & 0.353 & 0.877 & 0.550 & 1.476 & 0.346 \\ 
  &                     & AVG & 0.590 & 0.228 & 0.607 & 0.276 & \underline{0.822} & \underline{0.552} & 2.064 & \underline{0.377} & 0.744 & 0.465 & 1.348 & \underline{0.377} & 0.805 & 0.445 & 1.058 & \underline{0.383} \\ 
  \cmidrule(lr){2-19}
  & \multirow{2}{*}{5} & EN  & \textbf{0.669} & \textbf{0.290} & \textbf{0.856} & \textbf{0.254} & 0.872 & 0.696 & \textbf{3.215} & 0.353 & \textbf{0.806} & \textbf{0.576} & \textbf{2.186} & \textbf{0.359} & \textbf{0.882} & \textbf{0.571} & \textbf{1.523} & \textbf{0.349} \\ 
  &                    & AVG & \underline{0.593} & \underline{0.229} & \underline{0.616} & \underline{0.278} & 0.811 & 0.534 & 2.011 & 0.373 & \underline{0.752} & \underline{0.470} & 1.373 & 0.371 & \underline{0.811} & \underline{0.455} & \underline{1.081} & \underline{0.383} \\

\midrule
\multirow{6}{*}{GemmaVLM}
  & \multirow{2}{*}{3} & EN  & 0.613 & 0.245 & 0.578 & 0.220 & \textbf{0.879} & 0.697 & \textbf{3.221} & \textbf{0.355} & 0.769 & 0.531 & 2.066 & \textbf{0.356} & 0.851 & 0.505 & 1.351 & 0.337 \\ 
                          &                        & AVG & 0.562 & 0.194 & 0.531 & 0.267 & 0.811 & 0.535 & \underline{2.034} & 0.366 & 0.741 & 0.462 & \underline{1.383} & \underline{0.370} & 0.774 & 0.384 & 0.925 & 0.368     \\
  \cmidrule(lr){2-19}
  & \multirow{2}{*}{4} & EN  & 0.623 & 0.247 & 0.624 & 0.225 & \textbf{0.879} & \textbf{0.700} & 3.211 & 0.354 & 0.794 & 0.561 & 2.138 & \textbf{0.356} & 0.864 & 0.533 & 1.434 & 0.345 \\ 
  &                     & AVG & 0.568 & 0.203 & 0.550 & 0.268 & \underline{0.813} & \underline{0.536} & 1.975 & \underline{0.369} & 0.743 & \underline{0.463} & 1.330 & \underline{0.370} & 0.787 & 0.404 & 0.974 & 0.374 \\ 
  \cmidrule(lr){2-19}
  & \multirow{2}{*}{5} & EN  & \textbf{0.637} & \textbf{0.261} & \textbf{0.689} & \textbf{0.235} & 0.872 & 0.696 & 3.215 & 0.353 & \textbf{0.797} & \textbf{0.578} & \textbf{2.181} & 0.352 & \textbf{0.871} & \textbf{0.554} & \textbf{1.476} & \textbf{0.349} \\ 
  &                    & AVG & \underline{0.575} & \underline{0.207} & \underline{0.561} & \underline{0.271} & 0.795 & 0.511 & 1.880 & 0.362 & \underline{0.754} & 0.460 & 1.329 & 0.368 & \underline{0.795} & \underline{0.415} & \underline{0.997} & \underline{0.375} \\
\bottomrule
\end{tabular}
}
\end{table*}

\begin{figure}[!t]
    \centering

    \subfloat{
        \includegraphics[width=\linewidth]{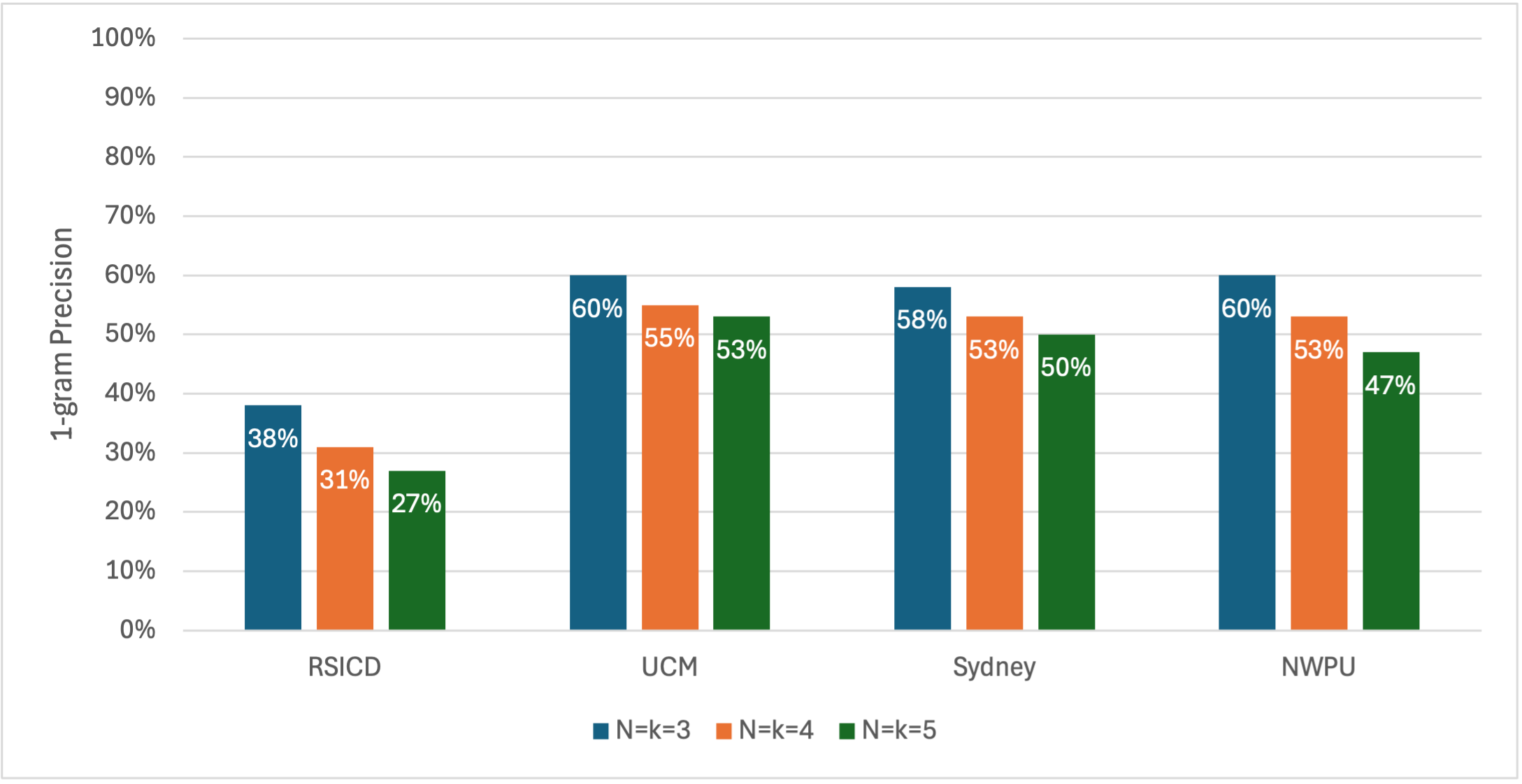}
    }\\[0.3cm] 

    \subfloat{
        \includegraphics[width=\linewidth]{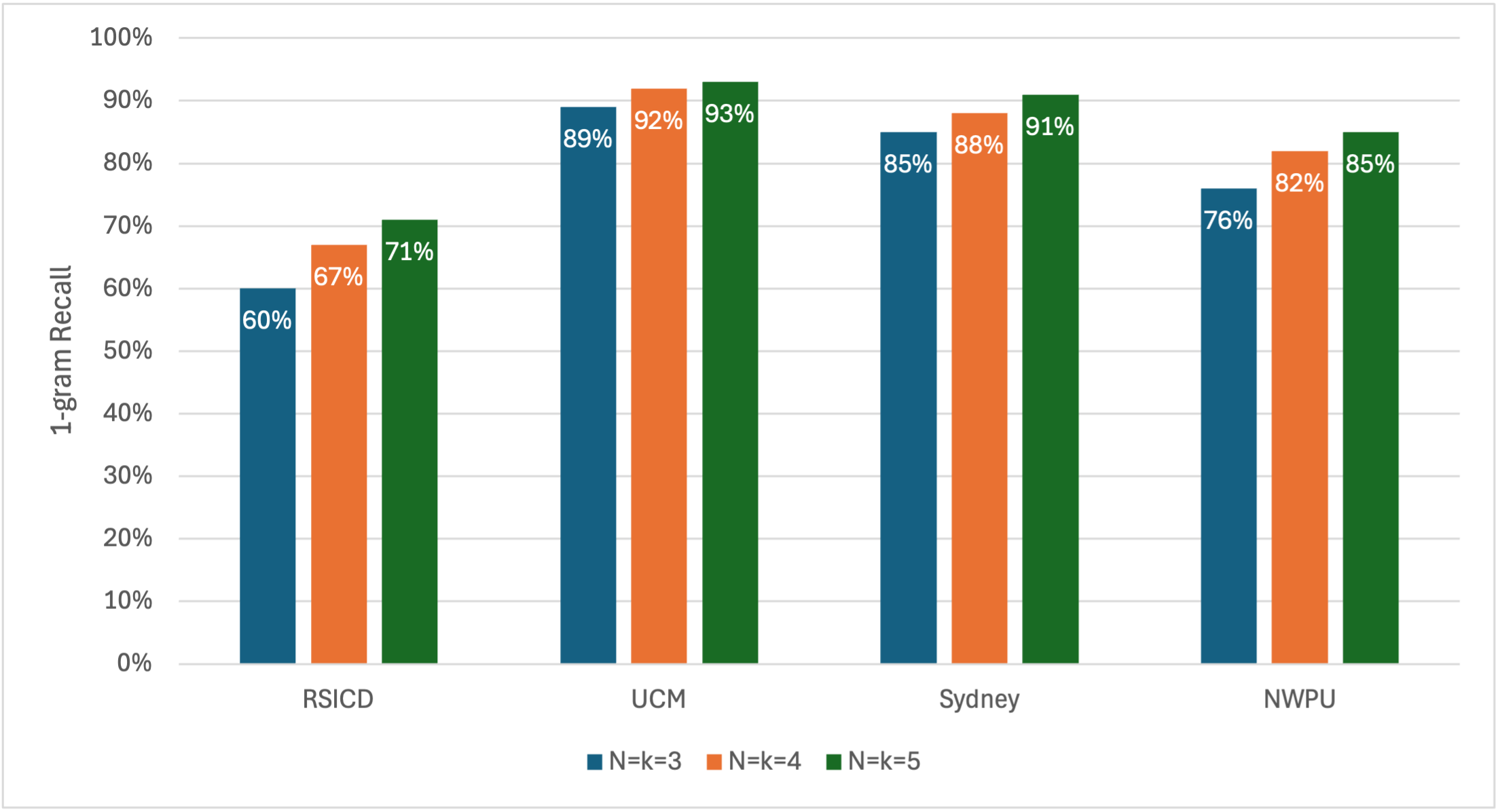}
    }

    \caption{Prompt-reference overlap (\%) for the three $N$ and $k$ values that were considered ($N$ = $k$ = 3, 4 and 5). The top chart reports the $1-$gram precision, and the bottom chart reports the $1-$gram recall.}
    \label{fig:prompt-ref-n-k}
\end{figure}

Figure \ref{fig:prompt-ref-n-k} presents the comparison of 1-gram precision and recall across the three configurations ($N$ = $k$ = 3, 4, 5). As expected, increasing the number of retrieved captions and few-shot examples leads to a decrease in precision, possibly due to the larger number of $n$-grams introduced into the prompt. However, this expansion also results in a clear and consistent increase in recall, showing that a greater proportion of reference $n$-grams is represented in the prompt.

This rise in recall provides a direct explanation for the performance gains observed in Table \ref{tab:ablation_N_K}. By exposing the model to more reference-aligned lexical and semantic content, the prompt becomes more informative and better grounded, enabling the model to produce captions that more closely resemble the reference descriptions.

\subsubsection{Image-aware Approach Without Retrieved Information}
\label{sec:no_retrieved_info}

\begin{figure}[!t]
  \includegraphics[width=\linewidth]{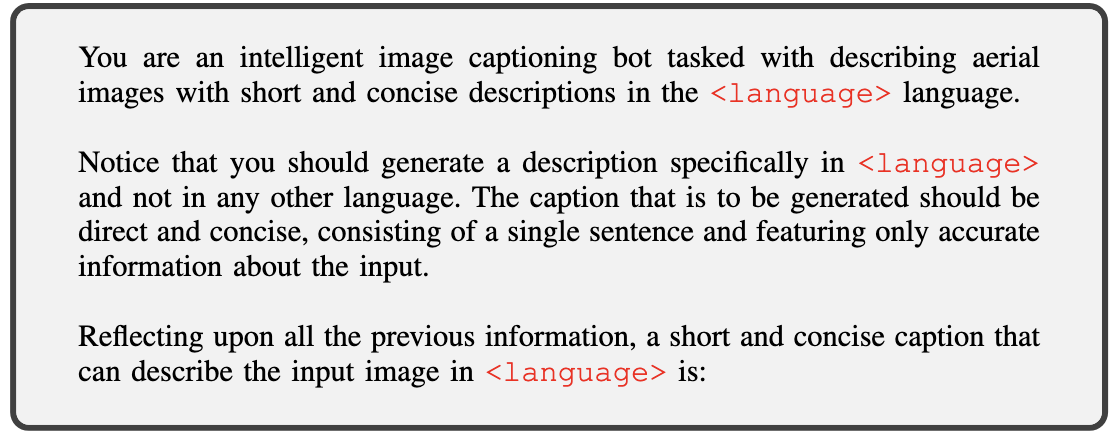}
\caption[Prompt given to the VLM without the use of retrieved captions or few-shot examples.]{Prompt given to the VLM for image captioning, without the use of retrieved captions or few-shot examples.}
\label{fig:simpleprompt}
\end{figure}

To assess the contribution of in-context learning to VLM performance, we conducted an additional experiment where models were prompted without retrieved captions or few-shot examples. This setup aimed to establish a baseline for the intrinsic captioning ability of the considered VLMs, isolating them from external guidance. The prompt, shown in Figure~\ref{fig:simpleprompt}, used the same initial instruction as the original, but excluded retrieved captions and few-shot examples.

Results in Table~\ref{tab:vlm_retrieval_comparison} show a consistent drop across all metrics, compared to the original strategy. Captions generated without in-context learning deviate from the expected style and structure, often omitting key details and failing to use terms that improve overlap with the ground-truth.

\begin{table*}
\centering
\caption{Comparison on the use of VLMs with and without few-shot examples and retrieved captions.}
\label{tab:vlm_retrieval_comparison}
\resizebox{\linewidth}{!}{
\begin{tabular}{lllcccccccccccccccc}
\toprule
\multirow{2}{*}{} & \multirow{2}{*}{} & \multirow{2}{*}{} & \multicolumn{4}{c}{RSICD} & \multicolumn{4}{c}{UCM} & \multicolumn{4}{c}{Sydney} & \multicolumn{4}{c}{NWPU} \\
\cmidrule(lr){4-7} \cmidrule(lr){8-11} \cmidrule(lr){12-15} \cmidrule(lr){16-19}
Model & Retrieval & Lang & BLEU1 & BLEU4 & CIDEr & SigLIP & BLEU1 & BLEU4 & CIDEr & SigLIP & BLEU1 & BLEU4 & CIDEr & SigLIP & BLEU1 & BLEU4 & CIDEr & SigLIP \\
\midrule
\multirow{4}{*}{EuroVLM} 
& \multirow{2}{*}{No}  & EN & 0.324 & 0.027 & 0.084 & 0.154 & 0.293 & 0.043 & 0.198 & 0.132 & 0.233 & 0.015 & 0.051 & 0.108 & 0.371 & 0.029 & 0.102 & 0.125 \\
&     & AVG & 0.274 & 0.034 & 0.056 & 0.171 & 0.286 & 0.041 & 0.104 & 0.125 & 0.296 & 0.040 & 0.066 & 0.150 & 0.325 & 0.041 & 0.063 & 0.150 \\
\cmidrule(lr){2-19}
& \multirow{2}{*}{Yes} & EN  & 0.571 & 0.213 & 0.605 & 0.225 & 0.870 & 0.687 & 3.435 & 0.360 & 0.768 & 0.555 & 2.295 & 0.360 & 0.788 & 0.479 & 1.321 & 0.351 \\
&     & AVG & 0.547 & 0.190 & 0.525 & 0.254 & 0.812 & 0.578 & 2.369 & 0.373 & 0.705 & 0.462 & 1.540 & 0.373 & 0.748 & 0.433 & 1.033 & 0.372 \\
\midrule
\multirow{4}{*}{GemmaVLM} 
& \multirow{2}{*}{No}  & EN  & 0.389 & 0.029 & 0.165 & 0.144 & 0.383 & 0.044 & 0.311 & 0.138 & 0.368 & 0.033 & 0.132 & 0.109 & 0.428 & 0.037 & 0.202 & 0.092 \\
&     & AVG & 0.345 & 0.039 & 0.124 & 0.153 & 0.348 & 0.052 & 0.205 & 0.127 & 0.375 & 0.050 & 0.106 & 0.138 & 0.392 & 0.043 & 0.134 & 0.142 \\
\cmidrule(lr){2-19}
& \multirow{2}{*}{Yes} & EN  & 0.613 & 0.245 & 0.578 & 0.220 & 0.879 & 0.697 & 3.221 & 0.355 & 0.769 & 0.531 & 2.066 & 0.356 & 0.851 & 0.505 & 1.351 & 0.337 \\
&     & AVG & 0.562 & 0.194 & 0.531 & 0.267 & 0.811 & 0.535 & 2.034 & 0.366 & 0.741 & 0.462 & 1.383 & 0.370 & 0.774 & 0.384 & 0.925 & 0.368 \\
\bottomrule
\end{tabular}
}
\end{table*}

\begin{figure}[!t]
  \centering
  \includegraphics[width=.95\linewidth]{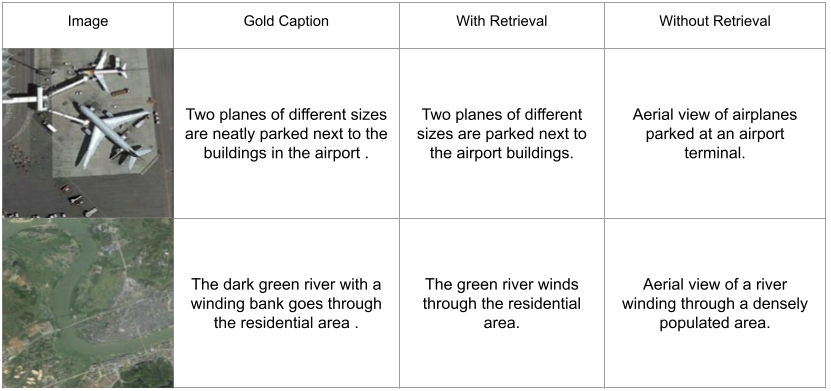}
  \caption[Example of generated captions from Gemma3-12B-VLM with and without few-shot examples and retrieved captions.]{Example illustrating the generated captions with the Gemma3-12B VLM, with and without few-shot examples and retrieved captions.}
  \label{fig:example_without_retrieval}
\end{figure}

The decline in RefSigLIPScore values is especially notable. One might expect this metric to remain stable, since captions are generated with access to the input image. However, without contextual examples, generalist VLMs produce captions that are stylistically and semantically misaligned with the remote sensing domain. Figure~\ref{fig:example_without_retrieval} illustrates this effect with a qualitative example from Gemma3-VLM, where captions without retrieval are simplified and less aligned than those from the original strategy.

\section{Conclusions}
\label{sec:conclusion}
This work advanced the understanding of training-free multilingual captioning methods for remote sensing imagery, by systematically comparing different models in two settings, namely an image-blind approach where LLMs are prompted without access to the input image, and an image-aware approach where VLMs are guided by both textual instructions and the image itself.

We achieved competitive results against state-of-the-art approaches that include supervised training, with the results obtained with the TowerInstruct-7B and Gemma3-12B-LM models standing out in particular.

Performance was mostly consistent across datasets, with RSICD standing as the main exception. Its limited lexical diversity may explain a reduced prompt–reference alignment, leading to weaker outputs. Comparing the image-blind and image-aware setups further revealed that, while VLMs often underperform on $n$-gram–based quality metrics due to generating visually grounded but unseen tokens, they remain competitive under the RefSigLIPScore metric, confirming semantic alignment with the input image.

Another key achievement was the assessment of captioning performance in nine languages beyond English, with generally strong results, despite the lack of human-annotated multilingual datasets. Crucially, we validated that directly prompting models in the target language yields better results than translating English captions post-hoc.

We also demonstrated the central role of retrieval quality. Fine-tuning the visual encoder significantly improved both caption selection and few-shot example matching, which translated into stronger prompt–reference alignment and higher-quality generations. The proposed PageRank-based re-ranking method further refined the retrieved content, boosting coherence and consistency with improvements of up to 35\%.

Extending the analysis, we showed that increasing the number of retrieved captions and few-shot examples can further strengthen generation quality. Larger $N$ and $k$ values enriched the prompt with more reference-aligned lexical and semantic content, yielding measurable gains across models.

Finally, we evaluated a baseline where VLMs were prompted without retrieved captions or few-shot examples. Performance dropped sharply, underscoring the decisive role of retrieval and prompt design in shaping both the style and semantic accuracy of the generated captions.

Several promising directions also arise from our experimental findings. A key priority is the creation of human-annotated multilingual captioning datasets, which would enable more reliable evaluation and reduce the limitations of translation-based references. Exploring diverse multimodal encoders, with different architectures, also appears essential, given the central role of retrieval quality.


Finally, future research should investigate the use of advanced reasoning models~\cite{deepseekai2025}, which may provide richer interpretative abilities (e.g., for assessing inconsistencies between all the retrieved information) and lead to the generation of more contextually accurate captions.

\section*{Acknowledgments}
This research was supported by the Portuguese Recovery and Resilience Plan through project C645008882-00000055 (i.e., the Center For Responsible AI), and also by the Fundação para a Ciência e Tecnologia (FCT), through the projects UIDB/50021/2020 and UIDP/04516/2020 (DOIs: \url{10.54499/UIDB/50021/2020}, \url{10.54499/UIDB/04516/2020}).


%
\bibliographystyle{IEEEtran}

\bibliography{bibliography}




\end{document}